\documentclass[10pt,twocolumn,letterpaper]{article}

\usepackage[pagenumbers]{cvpr} 

\definecolor{cvprblue}{rgb}{0.21,0.49,0.74}
\usepackage[pagebackref,breaklinks,colorlinks,allcolors=cvprblue]{hyperref}
\usepackage{cuted}
\usepackage{caption}
\usepackage[table]{xcolor}
\usepackage{booktabs}
\usepackage{placeins}
\usepackage{float}
\usepackage{mathtools}
\usepackage{wrapfig}
\usepackage{multirow}
\usepackage{tabularx}
\usepackage{etoc}
\usepackage{graphicx}
\usepackage{booktabs}
\usepackage{balance}
\usepackage[T1]{fontenc}
\usepackage[utf8]{inputenc}

\definecolor{specRed}{RGB}{162, 65, 53}
\colorlet{specRedLight}{specRed!25}

\definecolor{dinoBlue}{RGB}{63,76,100}
\colorlet{dinoBlueLight}{dinoBlue!25}

\definecolor{gainGreen}{HTML}{527D68}

\def\confName{CVPR}
\def\confYear{2027}

\title{Spec\textsuperscript{2}R: Recovering Emergent Properties in Supervised Vision Transformers}

\author{
Adam Wróbel$^{*1}$ \qquad
Jacek Tabor$^1$ \qquad
Bartosz Zieliński$^1$ \qquad
Marcin Przewi\k{e}źlikowski$^{1,2}$\\
\small$^1$Jagiellonian University, Kraków, Poland\qquad
\small$^2$NASK National Research Institute, Warsaw, Poland\\
{\tt\small
\{adam.wrobel,jacek.tabor,bartosz.zielinski,marcin.przewiezlikowski\}@uj.edu.pl}
}
\vspace{-5mm}

\begin{document}
\maketitle
\begin{strip}
    \centering
    \includegraphics[width=\textwidth]{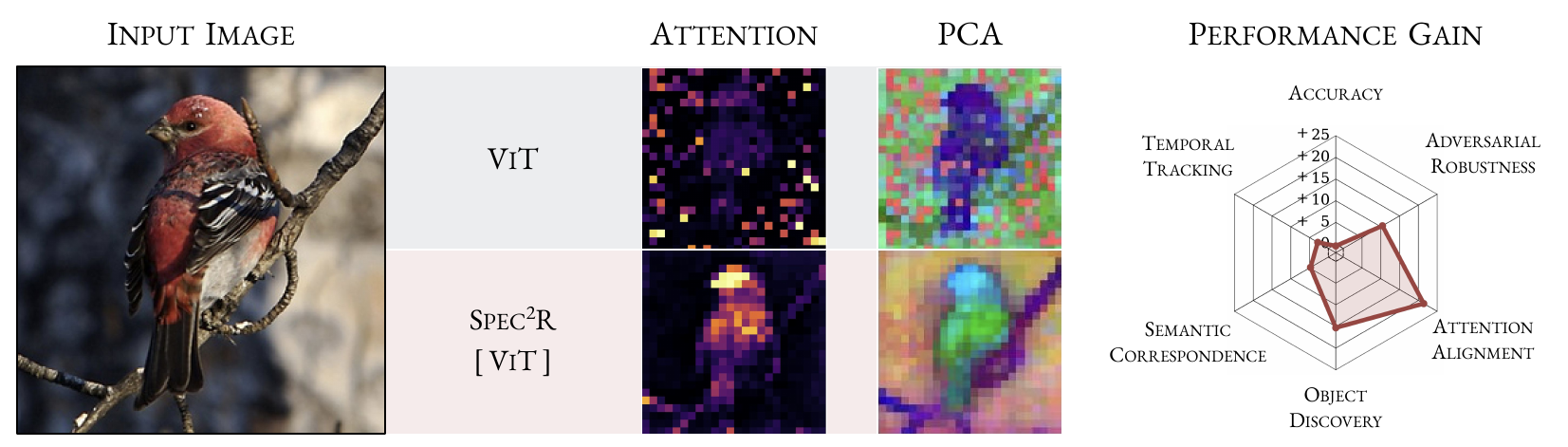}
    \captionof{figure}{
    \textbf{Effect of Spec\textsuperscript{2}R regularization on DeiT-III-B/16~\cite{touvron2022deit}.}
    By fine-tuning \textbf{only the MLP layers and classification head of a pretrained model}, Spec\textsuperscript{2}R improves object alignment in both patch representations and attention maps.
    These improvements translate into better downstream performance and adversarial robustness while preserving competitive classification accuracy.
    }
    \label{fig:teaser}
\end{strip}

\begin{abstract}
Vision Transformers (ViTs) have achieved strong performance across computer vision tasks, but the quality of their learned representations differs drastically depending on the training paradigm.
While self-supervised ViTs often demonstrate emergent properties, such as object-aligned patch representations and attention maps, supervised ViTs typically lack these properties, even when trained with strong data augmentations.
Understanding the origin of this difference is important for explaining and controlling the object-level structure.
To investigate it, we perform a spectral analysis of supervised and self-supervised models,
and identify a key distinction: MLP layers in supervised ViTs introduce high-frequency variations into patch representations, disrupting their spatial organization.
Motivated by this finding, we propose $\mathbf{Spec}^{2}\mathbf{R}$, a lightweight spectral regularization method that fine-tunes the MLP layers of pretrained supervised ViTs.
$\mathbf{Spec}^{2}\mathbf{R}$ recovers object-aligned patch representations and attention maps, while improving adversarial robustness and preserving competitive classification accuracy.
\end{abstract}
    
\section{Introduction}
\label{sec:intro}

Vision Transformers (ViTs)~\cite{dosovitskiy2020image} have become a widely adopted architecture for visual representation learning~\cite{khan2022transformers}. 
ViTs represent images as sequences of patch tokens, which interact through self-attention layers~\cite{vaswani2017attention}. 
As these representations evolve across transformer blocks, patch tokens can develop semantic structure, forming object-aligned clusters in the final representation~\cite{caron2021emerging, balestriero2025lejepa,przewiezlikowski2025cls}. 
Such ViT properties are particularly interesting because they often emerge without any explicit supervision at the level of individual image patches.

Self-supervised ViTs such as DINO~\cite{caron2021emerging, oquab2023dinov2, simeoni2025dinov3} provide a clear example, with object-aligned patch representations and attention maps that directly support correspondence~\cite{min2019spair, ham2017proposal}, tracking~\cite{pont20172017, ding2023mose}, and object discovery~\cite{wang2022self}.
Supervised ViTs~\cite{dosovitskiy2020image, touvron2022deit, touvron2021training, beyer2023flexivit}, despite their strong classification performance, show much weaker object-level structure.
This raises a question: \textit{What limits the emergence of object-aligned representations in supervised ViTs, and can these properties be recovered?}

\begin{figure*}[!t]
\includegraphics[width=\textwidth]{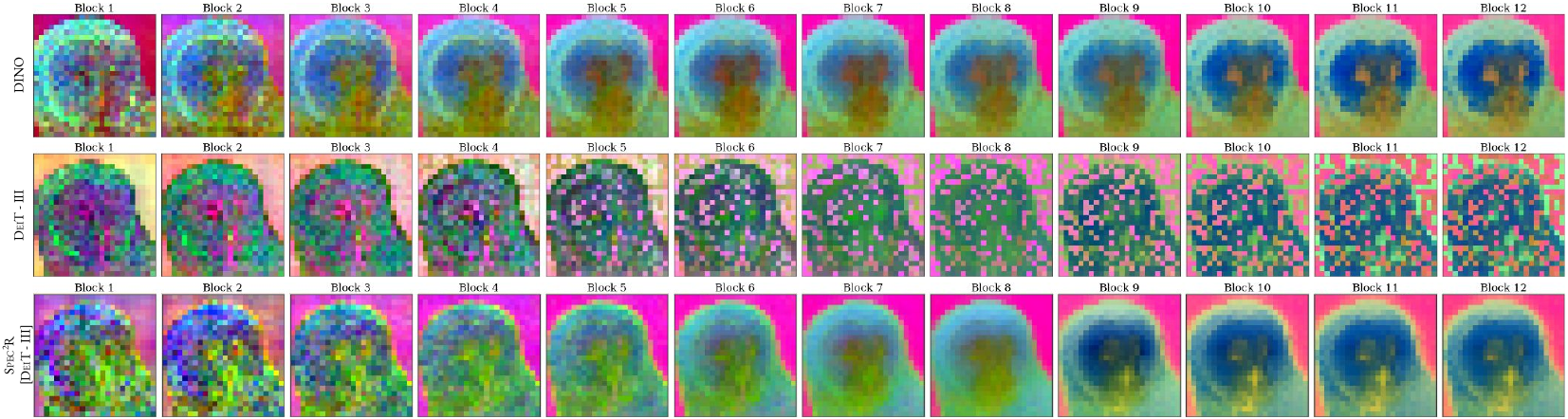}
    \caption{
        \textbf{Evolution of spatial structure across ViT blocks.}
        Patch representations from all 12 blocks are visualized as RGB projections onto the top three principal components of the final layer activation.
        DINO~\cite{caron2021emerging} (top) refines object-aligned structure with each layer, while DeiT-III~\cite{touvron2022deit} (middle) progressively introduces high-frequency spatial variations.
        Spec\textsuperscript{2}R[DeiT-III] (bottom) restores object-aligned structure through MLP and classification head fine-tuning.
    }
    \label{fig:analysis}
\end{figure*}

To investigate it, we use Fourier analysis to follow how the spatial frequency content of patch representations evolves across the blocks of DINO and supervised DeiT-III~\cite{touvron2022deit}, when projected onto dominant directions of final patch representations (Figure~\ref{fig:analysis}).
The two models follow opposite trajectories: DINO progressively refines its representations into object-aligned structure, whereas DeiT-III accumulates high-frequency variations that disrupt it.
We trace this difference to the MLP layers of supervised ViTs.

Motivated by this finding, we introduce Spec\textsuperscript{2}R, a lightweight spectral regularization that discourages the growth of high spatial frequencies across blocks by fine-tuning only the MLP layers and classification head of a pretrained supervised ViT.
Although all attention parameters remain frozen, Spec\textsuperscript2R improves the object alignment of attention maps, with DeiT-III and DeiT-Distilled~\cite{touvron2021training} surpassing DINO at the final block (Figure~\ref{fig:teaser}).
Together with consistent gains on downstream tasks and in adversarial robustness, these results suggest that object-aligned structure is neither exclusive to self-supervised training nor at odds with image-level classification.
Instead, it depends on how spatial frequencies evolve across transformer blocks, and can be recovered in pretrained supervised ViTs within a small fraction of a single ImageNet epoch.
\vspace{-4mm}
\paragraph{Contributions.}
Our main contributions are:
\begin{itemize}
\item A spectral analysis of ViT blocks, identifying MLP layers as a major source of excessive high frequencies in supervised ViTs.
\item Spec\textsuperscript{2}R, a lightweight spectral regularization that fine-tunes only MLP layers and the classification head.
\item Consistent improvements in attention alignment, patch representations, and adversarial robustness across supervised ViTs, with competitive classification accuracy.
\end{itemize}

\section{Related Works}
\label{sec:related_works}

\paragraph{ViT Representations.}
The original ViT~\cite{dosovitskiy2020image} demonstrated strong image classification performance through large-scale supervised pretraining.
DeiT~\cite{touvron2021training} introduced data-efficient training and CNN-based knowledge distillation, while DeiT-III~\cite{touvron2022deit} introduced a simplified augmentation-based training recipe.
FlexiViT~\cite{beyer2023flexivit} extended training to variable patch sizes, enabling flexible spatial tokenization.
Despite these advances, supervised ViTs often show weaker object alignment and lower performance on patch-level downstream tasks than self-supervised models.
DINO~\cite{caron2021emerging} introduced self-distillation for learning visual representations without labels, achieving emergent object-level structure in patch representations and attention maps despite the absence of explicit patch-level supervision.
DINOv2~\cite{oquab2023dinov2} combined image-level and patch-level self-distillation with large-scale training to produce general-purpose visual features.
DINOv3~\cite{simeoni2025dinov3} introduced Gram anchoring to preserve dense feature quality during extended training.
In this work, we use the original DINO as our self-supervised reference, as its object-level properties emerge without the additional training mechanisms used in later versions.

\paragraph{Understanding ViT Representations.}
Previous works have studied the mechanisms shaping ViT representations.
Wang et al.~\cite{wang2022anti} showed that self-attention tends to suppress high-frequency components of patch representations, potentially leading to oversmoothing in deep transformers.
Darcet et al.~\cite{darcet2024vision} identified high-norm tokens that emerge in low-information image regions.
Denoising Vision Transformers~\cite{yang2024denoising} attributed grid-like patterns in patch representations to positional embeddings structure.
More recently, Shi et al.~\cite{shi2026vision} investigated how global information aggregation can encourage background patches to encode object-level semantics.
Together, these works highlight several mechanisms affecting the spatial organization of ViT features.
Our analysis instead examines how individual transformer components affect the spatial frequency content of intermediate representations along the dominant directions of final patch features.
\vspace{-2mm}
\paragraph{Improving ViT Representations.}
Several approaches have managed to improve the spatial structure and semantic quality of ViT features.
SelfPatch~\cite{yun2022patch} encourages local semantic consistency through patch-level self-supervised learning.
Register tokens~\cite{darcet2024vision} provide additional capacity for internal information processing, reducing artifacts in dense feature maps.
SimPool~\cite{psomas2023keep} improves object alignment of attention maps through learnable attention-based pooling.
Denoising Vision Transformers~\cite{yang2024denoising} learns an additional network to remove positional artifacts from pretrained features, while ClearCLIP~\cite{lan2024clearclip} improves dense CLIP~\cite{radford2021learning} representations by modifying the contribution of residual features.
Unlike these approaches, Spec\textsuperscript{2}R directly refines pretrained supervised ViTs through lightweight MLP fine-tuning, without modifying attention parameters or introducing additional inference-time modules.
Our method constrains radial frequency expansion along dominant feature directions while preserving the model's classification behavior.
\section{Spectral Analysis}
\label{sec:analysis}
Our goal is to identify which transformer components may limit the emergence of object-aligned patch representations in supervised ViTs compared to self-supervised models such as DINO~\cite{caron2021emerging}.
Supervised ViTs are known to produce lower-quality attention maps than their self-supervised counterparts~\cite{caron2021emerging, psomas2023keep}, and both their attention maps and patch representations appear spatially noisy, with fine-grained variations that do not follow object boundaries (Figure~\ref{fig:teaser}).
Since such noise corresponds to high spatial frequencies, we use Fourier analysis to study the \emph{spectral dynamics} of patch representations, \ie, how their spatial frequency content evolves across transformer blocks.

\subsection{Preliminaries}
To provide the foundation for our analysis and the Spec\textsuperscript{2}R regularization (Section~\ref{sec:spec2r}), we first introduce the notation and measures used to characterize the dynamics of patch representations across transformer blocks.
Additional mathematical details are provided in Appendix~\ref{app:formulation}.

\paragraph{Residual stream.}
Our analysis focuses on intermediate patch representations produced by ViT blocks.
Let $V \cong \mathbb{R}^{t \times d}$ denote the space of patch representations, where $t$ is the number of patch tokens and $d$ is the embedding dimension.
Each ViT block $F_i$ consists of a residual attention branch $F_i^{\mathrm{attn}}$ followed by a residual MLP branch $F_i^{\mathrm{mlp}}$, such that
$F_i = F_i^{\mathrm{mlp}} \circ F_i^{\mathrm{attn}}$.
Given an initial representation $\boldsymbol{X}_0 \in V$, a sequence of $k$ blocks produces
$\boldsymbol{X}_{i+1} = F_i(\boldsymbol{X}_i)$.
We refer to the sequence $(\boldsymbol{X}_i)_{i=0}^{k}$ as the \emph{residual stream}.

\paragraph{Spectral measures.}
To characterize changes in spatial frequencies across the residual stream, we consider the radial and angular spectra of the two-dimensional Fourier amplitude spectrum.
Let $\mathcal{F}_{2D}(\boldsymbol{Z})$ denote the 2D Fourier transform of a spatial representation $\boldsymbol{Z}$, defined over the frequency domain $\Omega \subset \mathbb{R}^2$.
For any measurable subset $S \subseteq \Omega$, we define its accumulated amplitude spectrum as:
\begin{equation}
    \Phi(\boldsymbol{Z}, S) \coloneqq \int_S |\mathcal{F}_{2D}(\boldsymbol{Z})(\boldsymbol{\omega})|\,d\boldsymbol{\omega}
\end{equation}
\noindent
The radial and angular spectra are then defined by choosing appropriate interval sets:
\begin{equation}
\begin{alignedat}{2}
\Phi_r(\boldsymbol{Z}, B)
&\coloneqq \Phi(\boldsymbol{Z},
\{\boldsymbol{\omega}\in\Omega:
\|\boldsymbol{\omega}\|_2 &\in B\})\\
\Phi_{\theta}(\boldsymbol{Z}, B)
&\coloneqq \Phi(\boldsymbol{Z},
\{\boldsymbol{\omega}\in\Omega:
\arg(\boldsymbol{\omega}) &\in B\})
\end{alignedat}
\label{eq:spectra}
\end{equation}

\noindent 
where $B$ denotes radial or angular frequency intervals, respectively. The radial spectrum $\Phi_r$ quantifies how spectral amplitude is distributed between low and high spatial frequencies, while the angular spectrum $\Phi_{\theta}$ quantifies its distribution across orientations, providing complementary information.

\subsection{Spectral Dynamics}
Building on the spectral measures, we analyze how the residual stream $(\boldsymbol{X}_i)_{i=0}^k$ evolves along the dominant directions of the final layer patch representations $\boldsymbol{X}_k$.
We compare self-supervised DINO with supervised DeiT-III, \textbf{with consistent observations across additional architectures reported in the Appendix}~\ref{app:spectral_analysis}.

\paragraph{Residual stream projection.}
For each image, we identify the dominant feature directions of the final-layer patch representations $\boldsymbol{X}_k$ through PCA, selecting the three principal components associated with the largest singular values.
These components form a reference basis $\mathcal{B}_k$.
We then project each intermediate representation $\boldsymbol{X}_i$ onto this basis, obtaining the projected residual stream $([\boldsymbol{X}_i]_{\mathcal{B}_k})_{i=0}^k$.
As visualized in Figure~\ref{fig:analysis}, while DINO projection progressively refines object-aligned structure, DeiT-III introduces high-frequency variations at block 5, progressively disrupting its spatial representation along $\mathcal{B}_k$.

\paragraph{Radial Moment Trajectory.}
\begin{figure}[!t]
    \includegraphics[width=1.0\linewidth]{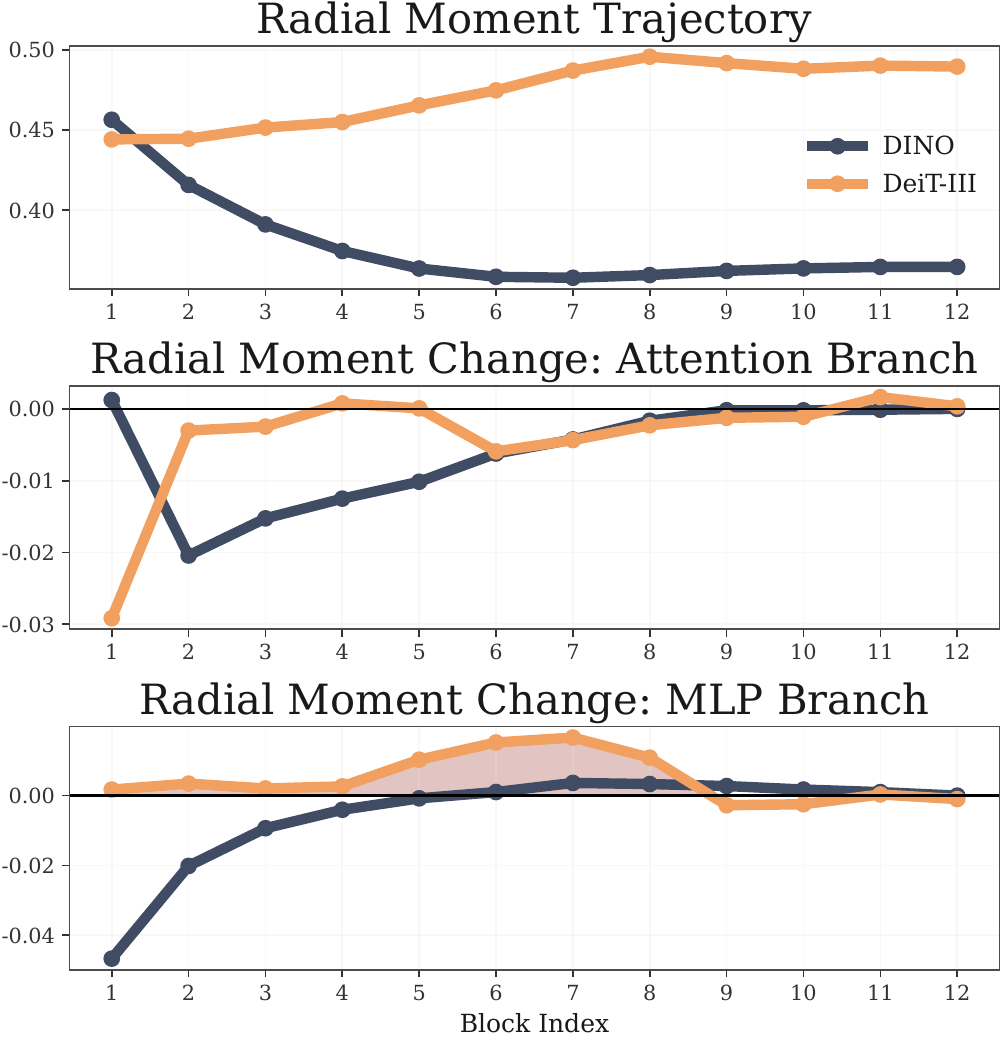}
    \caption{
    \textbf{Radial moment dynamics across ViT blocks.}
    Top: Radial moment trajectories averaged over the ImageNet-1K~\cite{imagenet} validation set.
    Middle and bottom: Changes in radial moment induced by MLP and attention residual branches ($F_i^{\mathrm{mlp}}$ and $F_i^{\mathrm{attn}}$).
    DeiT-III~\cite{touvron2022deit} MLPs increase the radial moment (red region), producing the accumulating spatial variations.
    }
    \label{fig:radial_spectre}
\end{figure}

To quantify these observations, we reshape each projected representation $[\boldsymbol{X}_i]_{\mathcal{B}_k}$ into a spatial grid of size $g \times g \times |\mathcal{B}_k|$ and compute the sum-normalized radial spectrum $\tilde{\Phi}_r$ independently for each PCA component.
We then summarize the frequency distribution at each block by the first radial moment, averaged across the reference directions:
\begin{equation}
\mu_r^{(i)} \coloneqq
\frac{1}{|\mathcal{B}_k|}
\sum_{b \in \mathcal{B}_k}
\mathbb{E}_{r \sim \tilde{\Phi}_r([\boldsymbol{X}_i]_b)}[r]
\label{eq:radial_moment}
\end{equation}
We refer to the resulting sequence $(\mu_r^{(i)})_{i=0}^k$ as the \emph{radial moment trajectory}, which describes the evolution of the expected radial frequency across transformer blocks.

As shown in Figure~\ref{fig:radial_spectre} (top), DINO shows a decreasing radial moment across early blocks (1-7), followed by an approximately non-increasing trend, whereas DeiT-III progressively shifts toward higher spatial frequencies.
These contrasting trajectories show that the two models differ not only in their final patch representations, but also in how their representations evolve across transformer blocks.
This observation is consistent with the qualitative differences in Figure~\ref{fig:analysis}, where DINO progressively refines spatial structure while DeiT-III develops high-frequency variations.

To identify which components of each ViT block $F_i$ contribute to changes in the radial moment trajectory, we measure the difference in radial moment before and after each residual branch ($F_i^{\mathrm{attn}}$ and $F_i^{\mathrm{mlp}}$).
As shown in Figure~\ref{fig:radial_spectre} (middle and bottom), attention branch contributions remain approximately non-positive in both DINO and DeiT-III.
In contrast, DeiT-III shows consistently positive MLP branch contributions in blocks 5-8, shifting the representations toward higher spatial frequencies.
We observe that similar MLP behavior repeats across other supervised ViTs (see Appendix~\ref{app:spectral_analysis}).

\paragraph{Angular similarity.}
\begin{figure}[!t]
    \includegraphics[width=1.0\linewidth]{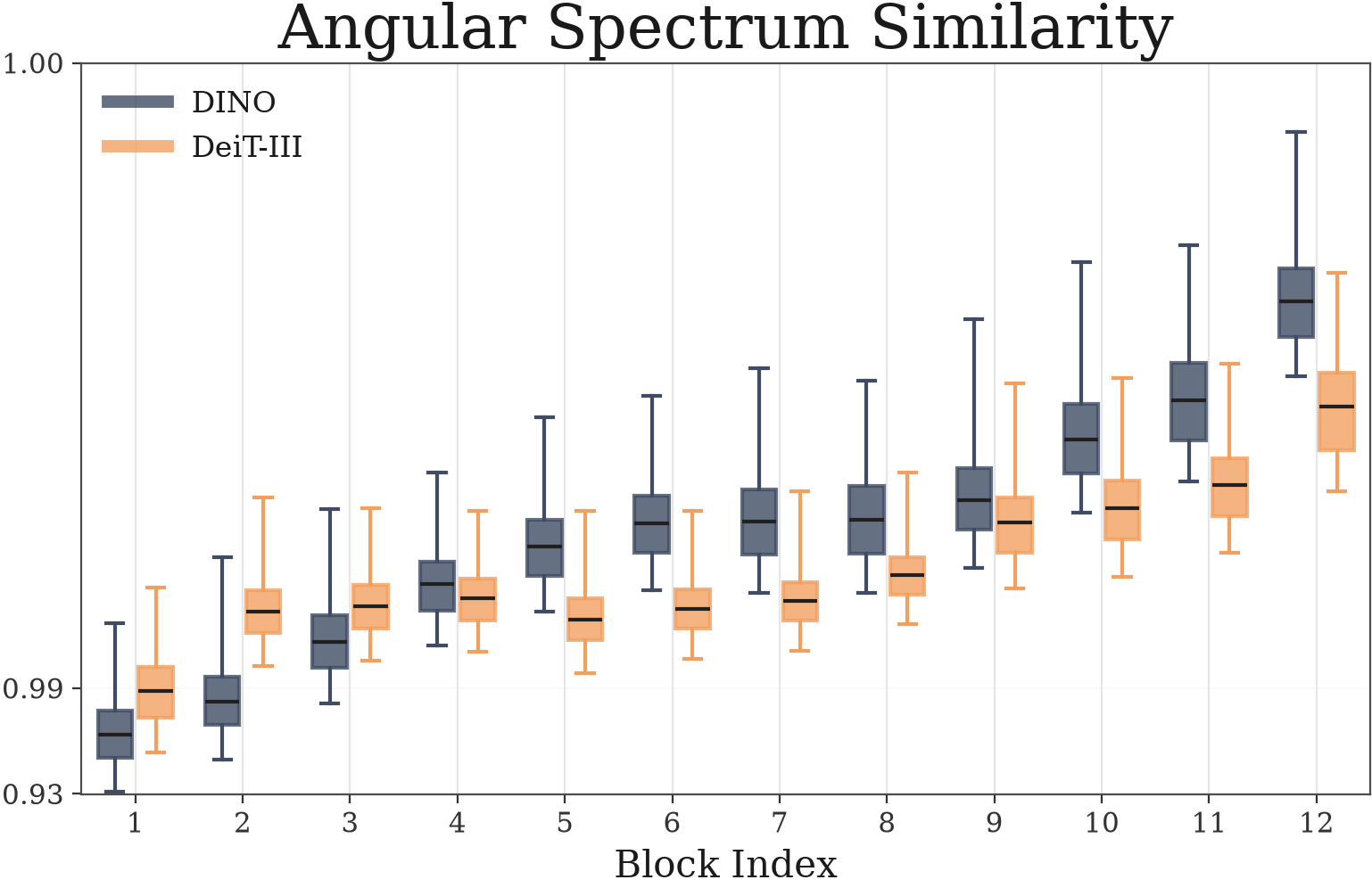}
    \caption{
    \textbf{Angular spectrum similarity across ViT blocks.}
    Box plots of cosine similarity between normalized angular spectra before and after each transformer block, for DINO~\cite{caron2021emerging} and DeiT-III~\cite{touvron2022deit}.
    Despite contrasting radial dynamics (see Figure~\ref{fig:radial_spectre}), both models maintain high angular similarity, motivating its preservation during spectral regularization.
    }
    \label{fig:angular_spectre}
\end{figure}
While the radial moment captures changes in spatial frequency magnitude, it does not describe how spectral amplitude is distributed across spatial orientations (see Equation~\ref{eq:spectra}).
To complement the radial analysis, we measure the cosine similarity between sum-normalized angular spectra $\tilde{\Phi}_{\theta}$ before and after each ViT block:
\begin{equation}
s_\theta^{(i)} \coloneqq
\frac{1}{|\mathcal{B}_k|}
\sum_{b \in \mathcal{B}_k}
\operatorname{cos}\left(
\tilde{\Phi}_\theta([\boldsymbol{X}_{i-1}]_b),
\tilde{\Phi}_\theta([\boldsymbol{X}_i]_b)
\right)
\label{eq:angular_similarity}
\end{equation}
This similarity quantifies the stability of the angular spectrum $\tilde{\Phi}_{\theta}$ along the projected residual stream $([\boldsymbol{X}_i]_{\mathcal{B}_k})_{i=0}^k$.

As shown in Figure~\ref{fig:angular_spectre}, both DINO and DeiT-III maintain high angular similarity between their blocks, starting above $0.93$ for $s_{\theta}^{(1)}$ and progressively increasing over $0.99$.
This shows that despite their contrasting radial dynamics, both models preserve their angular distribution of spectral amplitude between successive blocks, suggesting that this stability is a useful constraint when correcting radial frequency expansion.

\paragraph{Conclusions.}
Our analysis shows contrasting radial dynamics despite stable angular spectra in both DINO and DeiT-III.
While DINO progressively reduces the radial moment, DeiT-III increases it through high-frequency variations introduced by its MLP branches.
These findings are consistent across all considered ViTs (see Appendix~\ref{app:spectral_analysis}), and motivate our spectral regularization, which limits radial frequency expansion while preserving the angular similarity across successive transformer blocks.

\section{Spec\textsuperscript{2}R}
\label{sec:spec2r}
We introduce Spec\textsuperscript{2}R, a lightweight spectral regularization method that promotes DINO-like spatial organization in the patch representations of pretrained supervised ViTs.
Following our analysis from the previous section, \textbf{our fine-tuning is restricted to the MLP branches and classification head, while all remaining parameters remain frozen.}

\paragraph{Radial non-expansion.}
Motivated by the positive MLP contributions shown in our analysis, the first regularization term of Spec\textsuperscript{2}R penalizes increases in the radial moment across successive transformer blocks, independently for each PCA component:

\begin{equation}
\mathcal{L}_r \coloneqq
\frac{1}{|\mathcal{B}_k|}
\sum_{i=1}^{k} \sum_{b \in \mathcal{B}_k}
\left[
\mu_{r,b}^{(i)}
-
\mathrm{sg}[\mu_{r,b}^{(i-1)}]
\right]_+
\end{equation}
Here, $\mu_{r,b}^{(i)}$ denotes the first radial moment of the $b$-th projected component, and $\mathrm{sg}[\cdot]$ denotes the stop-gradient operator, which we also apply to per-image reference PCA basis $\mathcal{B}_k$ (see Equation~\ref{eq:radial_moment}).

\paragraph{Angular preservation.}
The second component of Spec\textsuperscript{2}R encourages the preservation of angular spectrum similarity across the projected residual stream.
Following the high angular similarity observed in both DINO and DeiT-III, we minimize the cosine distance between normalized angular spectra before and after each ViT block:
\begin{equation}
    \mathcal{L}_{\theta} \coloneqq
    \sum_{i=1}^{k}
    \left[
    1 - s_{\theta}^{(i)}\right]
\end{equation}
Similarly to the previous objective, we apply the stop gradient operator $\mathrm{sg}[.]$ on the reference basis $\mathcal{B}_k$ and on the angular spectrum of input activation $\tilde{\Phi}_{\theta}([\boldsymbol{X}_{i-1}]_b)$ (see Equation~\ref{eq:angular_similarity}).
This constraint discourages changes in the angular distribution of the spectral amplitude while allowing the radial moment trajectory to adjust. 
It also helps prevent the performance degradation observed with radial regularization alone, as demonstrated in Appendix~\ref{app:angular_ablation}.

\paragraph{Fine-tuning objective.}
To preserve the predictive behavior of the pretrained model, we use the Kullback-Leibler divergence~\cite{hinton2015distilling} between the output probability distributions of the fine-tuned and original models, denoted by $\mathcal{L}_{\mathrm{KL}}$.
The final fine-tuning objective combines the distillation term with the radial non-expansion and angular preservation losses:
\begin{equation}
\label{eq:radial_loss}
    \mathcal{L}_{\mathrm{Spec^2R}}
    \coloneqq
    \lambda_{KL}\mathcal{L}_{\mathrm{KL}} +
    \lambda_r \mathcal{L}_r
    + \lambda_\theta \mathcal{L}_\theta
\end{equation}
where $\lambda_{KL}$, $\lambda_r$ and $\lambda_\theta$ control the contributions of the loss terms.

\begin{figure}
    \centering
    \includegraphics[width=\linewidth]{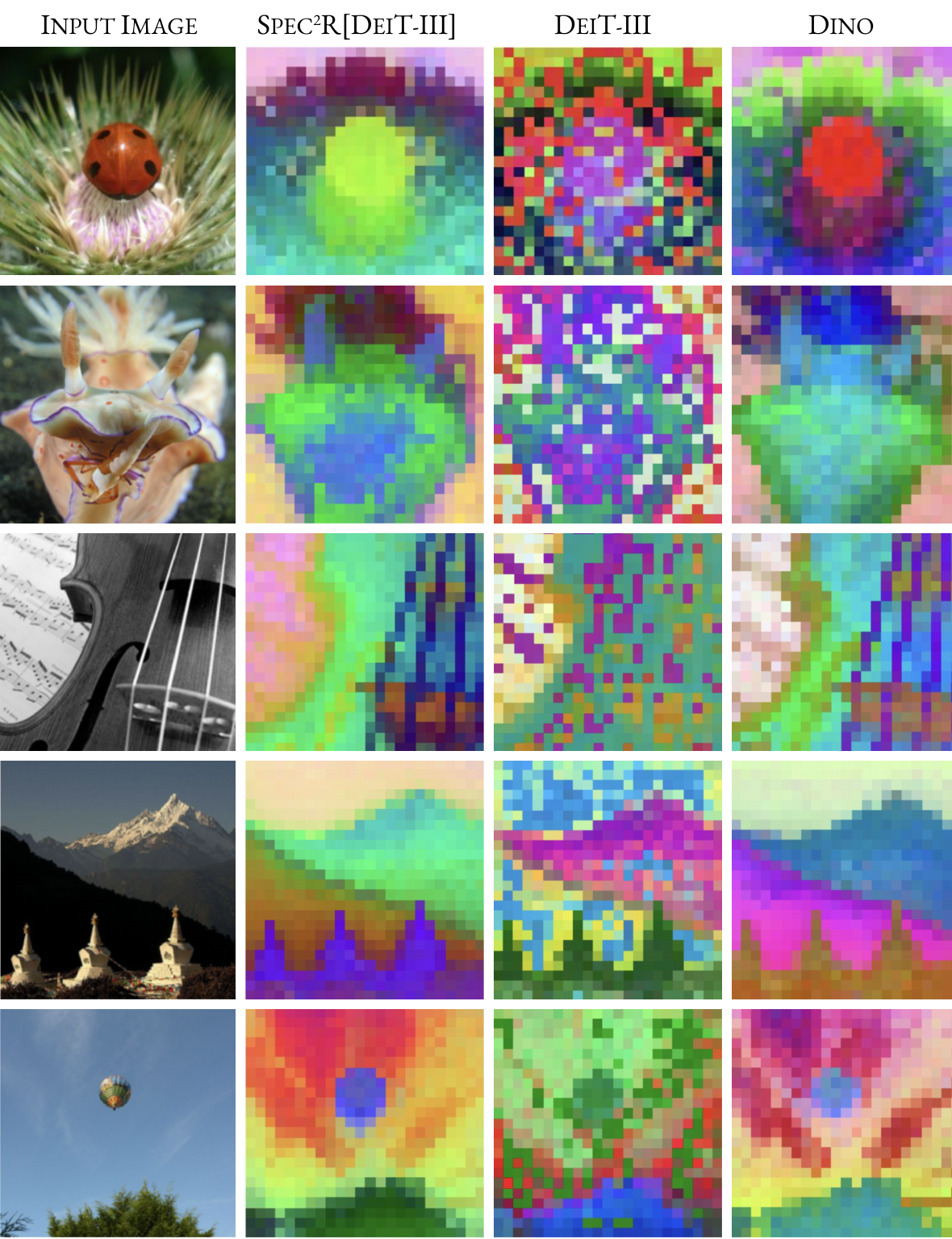}
    \caption{
    \textbf{Qualitative comparison of final-layer patch representations.}
    RGB visualizations of the top three principal components of patch tokens on ImageNet-1K~\cite{imagenet} images for DINO~\cite{caron2021emerging}, DeiT-III~\cite{touvron2022deit}, and Spec\textsuperscript{2}R [DeiT-III].
    Spec\textsuperscript{2}R removes high-frequency artifacts from DeiT-III representations and improves object alignment, producing representations similar to DINO.
}
    \label{fig:pca_gallery}
\end{figure}
\begin{figure*}[!t]
    \centering
    \includegraphics[width=\textwidth]{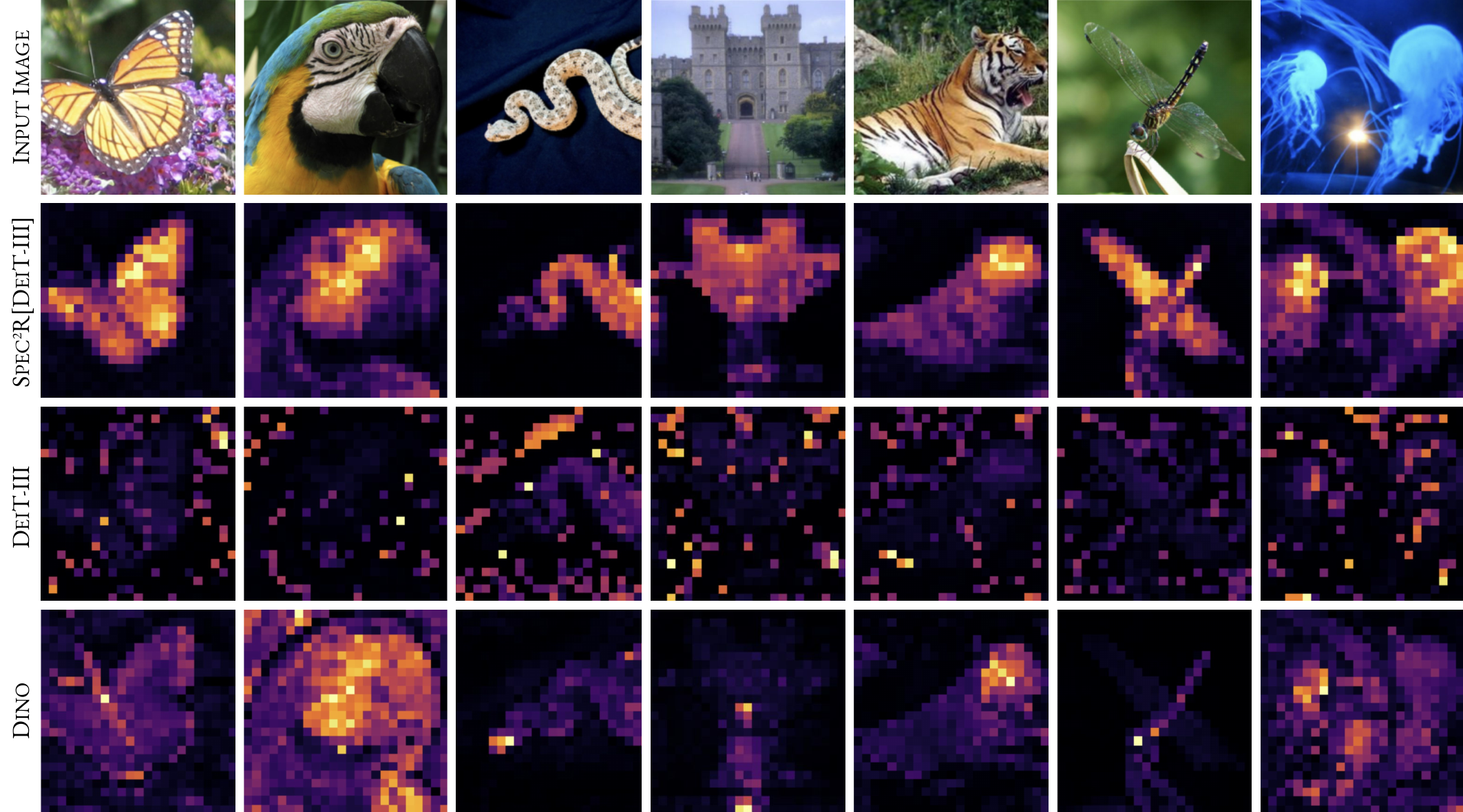}
\caption{
\textbf{Qualitative comparison of final-layer CLS-token attention.}
CLS-token attention maps averaged across heads on ImageNet-1K~\cite{imagenet} images for DINO~\cite{caron2021emerging}, DeiT-III~\cite{touvron2022deit}, and Spec\textsuperscript{2}R [DeiT-III].
Despite keeping attention parameters frozen, Spec\textsuperscript{2}R[DeiT-III] improves object-alignment of attention maps, allowing DeiT-III to produce more object-aligned attention maps than DINO.
}
    \label{fig:attention_gallery}
\end{figure*}

\section{Experiments}
\subsection{Experimental Setup}
\label{sec:experiments}

\paragraph{Models.}
To cover diverse supervised training strategies, we select three ViT-B/16 models: DeiT-III~\cite{touvron2022deit}, DeiT-Distilled~\cite{touvron2021training}, and FlexiViT~\cite{beyer2023flexivit}.
DeiT-III uses training recipe based on a simplified data augmentation pipeline. 
DeiT-Distilled uses knowledge distillation from a CNN teacher through an additional token.
FlexiViT uses varying patch sizes to support flexible spatial tokenization.
Additionally, we use self-supervised DINO~\cite{caron2021emerging} as a reference due to its analogous role in our analysis from Section~\ref{sec:analysis}.

\paragraph{Fine-tuning.}
We fine-tune only the MLP layers and classification head, keeping all remaining parameters frozen.
We use Adam optimizer~\cite{kingma2014adam} with a batch size of 64 and a two-stage optimization procedure.
The first stage runs for approximately 300 steps with a lower KL loss weight.
Once the losses plateau, we reset the optimizer and continue for approximately 1,000 steps with a reduced learning rate and increased KL loss weight to better preserve classification performance.
The full optimization schedule and model-specific hyperparameters are provided in the Appendix~\ref{app:finetuning}.

\paragraph{Evaluation.}
We evaluate patch representation quality across three complementary downstream tasks.
(i) Semantic correspondence measures the semantic consistency of patch representations across images.
(ii) Temporal tracking evaluates instance consistency of patch representations across video frames.
(iii) Object discovery assesses the ability to identify structured object regions from patch representations.
We also assess the object alignment of CLS-token attention maps, measure classification performance and adversarial robustness of fine-tuned models.

\begin{table}[t]
    \centering
    \caption{
        \textbf{Object alignment of CLS-token attention.}
        We evaluate the best attention head using Energy Pointing
        Game (EPG)~\cite{wang2020score} on ImageNet-1K~\cite{imagenet} and foreground intersection-over-union
        (IoU) on PASCAL VOC~\cite{everingham2010pascal}.
        We report performance at the final transformer block (Last), the highest performance across blocks (Best), and the corresponding block index (Best Block).
    }
    \scriptsize
    \setlength{\tabcolsep}{3.5pt}

    \begin{tabular}{lcccccc}
        \toprule
        & \multicolumn{3}{c}{\textbf{ImageNet-1K (EPG)}}
        & \multicolumn{3}{c}{\textbf{PASCAL VOC (IoU)}} \\
        \cmidrule(lr){2-4}
        \cmidrule(lr){5-7}
        Model
            & Last $\uparrow$ & Best $\uparrow$ & Best Block
            & Last $\uparrow$ & Best $\uparrow$ & Best Block \\
        \midrule
        \rowcolor{dinoBlueLight}
        DINO-B/16
            & 78.89 & 82.94 & 10
            & 30.95 & 39.20 & 6 \\
        \midrule
        
        DeiT-III-B/16
            & 66.09 & 78.19 & 9
            & 17.04 & 43.84 & 8 \\
        \rowcolor{specRedLight}
        + Spec$^2$R
            & \underline{\textbf{85.84}} & \underline{\textbf{85.84}} & 12
            & \underline{\textbf{40.16}} & \underline{\textbf{46.11}} & 8 \\

        $\Delta$ &
\textcolor{gainGreen}{\textbf{+19.75}} &
\textcolor{gainGreen}{\textbf{+7.65}} &
&
\textcolor{gainGreen}{\textbf{+23.12}} &
\textcolor{gainGreen}{\textbf{+2.27}} &  \\
        \midrule
        DeiT-Distilled-B/16
            & 64.15 & 76.89 & 9
            & 14.65 & 35.71 & 6 \\
        \rowcolor{specRedLight}
        + Spec$^2$R
            & \textbf{79.43} & \textbf{81.48} & 10
            & \textbf{33.26} & \textbf{36.63} & 6 \\
        $\Delta$ & \textcolor{gainGreen}{\textbf{+15.28}} & \textcolor{gainGreen}{\textbf{+4.59}} & & \textcolor{gainGreen}{\textbf{+18.61}}& \textcolor{gainGreen}{\textbf{+0.92}} &  \\
        \midrule
        FlexiViT-B/16
            & 65.43 & 73.79 & 10
            & 21.72 & 41.10 & 8 \\
        \rowcolor{specRedLight}
        + Spec$^2$R
            & \textbf{74.78} & \textbf{77.45} & 11
            & \textbf{37.60} & \textbf{42.93} & 8 \\
        $\Delta$ & \textcolor{gainGreen}{\textbf{+9.35}} & \textcolor{gainGreen}{\textbf{+3.66}} & & \textcolor{gainGreen}{\textbf{+15.88}}& \textcolor{gainGreen}{\textbf{+1.83}} &  \\
        \bottomrule
    \label{tab:attention_main}
    \end{tabular}
\vspace{-5mm}
\end{table}

\begin{table*}[t]
\centering
\caption{
\textbf{Evaluation of patch representations across downstream tasks.}
Semantic correspondence is evaluated using PCK~\cite{yang2012articulated} at three thresholds
on SPair-71k~\cite{min2019spair} and PF-PASCAL~\cite{ham2017proposal}.
Temporal tracking is evaluated on DAVIS~\cite{pont20172017} and MOSE~\cite{ding2023mose} using
$\mathcal{J}$, $\mathcal{F}$, and their mean.
Object discovery is evaluated using TokenCut~\cite{wang2022self} CorLoc on
PASCAL VOC07, VOC12~\cite{everingham2010pascal}, and COCO20K~\cite{lin2014microsoft}.
$\Delta$ denotes the absolute improvement over the corresponding
pretrained model in percentage points.
}
\label{tab:downstream}

\resizebox{\textwidth}{!}{%
\begin{tabular}{lccccccccccccccc}
\toprule
& \multicolumn{6}{c}{\textbf{Semantic Correspondence}}
& \multicolumn{6}{c}{\textbf{Temporal Tracking}}
& \multicolumn{3}{c}{\textbf{Object Discovery}} \\
\cmidrule(lr){2-7}
\cmidrule(lr){8-13}
\cmidrule(lr){14-16}

& \multicolumn{3}{c}{\textbf{SPair-71k}}
& \multicolumn{3}{c}{\textbf{PF-PASCAL}}
& \multicolumn{3}{c}{\textbf{DAVIS}}
& \multicolumn{3}{c}{\textbf{MOSE}}
& \multicolumn{3}{c}{\textbf{TokenCut}} \\
\cmidrule(lr){2-4}
\cmidrule(lr){5-7}
\cmidrule(lr){8-10}
\cmidrule(lr){11-13}
\cmidrule(lr){14-16}

Model
& 0.05 $\uparrow$ & 0.10 $\uparrow$ & 0.20 $\uparrow$
& 0.05 $\uparrow$ & 0.10 $\uparrow$ & 0.20 $\uparrow$
& $\mathcal{J}\uparrow$ & $\mathcal{F}\uparrow$
& $\mathcal{J}\&\mathcal{F}\uparrow$
& $\mathcal{J}\uparrow$ & $\mathcal{F}\uparrow$
& $\mathcal{J}\&\mathcal{F}\uparrow$
& VOC07 $\uparrow$ & VOC12 $\uparrow$ & COCO20K $\uparrow$ \\
\midrule

\rowcolor{dinoBlueLight}
DINO-B/16
& \underline{11.81} & 26.99 & 46.21
& 28.58 & 51.41 & 69.51
& \underline{60.69} & \underline{63.86} & \underline{62.28}
& 31.63 & \underline{38.43} & \underline{35.03}
& \underline{57.37} & \underline{62.54} & \underline{48.20} \\
\midrule

DeiT-Distilled-B/16
& 11.31 & 27.37 & 48.16
& 27.17 & 49.63 & 68.68
& 59.79 & 62.19 & 60.99
& 30.23 & 37.05 & 33.64
& 52.62 & 56.22 & 45.72 \\

\rowcolor{specRedLight}
+ Spec$^2$R
& \textbf{11.72} & \underline{\textbf{28.39}}
& \underline{\textbf{50.06}}
& \underline{\textbf{29.08}}
& \underline{\textbf{53.77}}
& \underline{\textbf{72.29}}
& \textbf{60.60} & \textbf{62.85} & \textbf{61.73}
& \textbf{31.67} & \textbf{38.36} & \textbf{35.01}
& \textbf{54.44} & \textbf{59.35} & \textbf{46.12} \\

$\Delta$
& \textcolor{gainGreen}{\textbf{+0.41}}
& \textcolor{gainGreen}{\textbf{+1.02}}
& \textcolor{gainGreen}{\textbf{+1.90}}
& \textcolor{gainGreen}{\textbf{+1.91}}
& \textcolor{gainGreen}{\textbf{+4.14}}
& \textcolor{gainGreen}{\textbf{+3.61}}
& \textcolor{gainGreen}{\textbf{+0.81}}
& \textcolor{gainGreen}{\textbf{+0.66}}
& \textcolor{gainGreen}{\textbf{+0.74}}
& \textcolor{gainGreen}{\textbf{+1.44}}
& \textcolor{gainGreen}{\textbf{+1.31}}
& \textcolor{gainGreen}{\textbf{+1.37}}
& \textcolor{gainGreen}{\textbf{+1.82}}
& \textcolor{gainGreen}{\textbf{+3.13}}
& \textcolor{gainGreen}{\textbf{+0.40}} \\
\midrule

DeiT-III-B/16
& 7.77 & 19.60 & 37.46
& 19.76 & 38.36 & 57.83
& 54.30 & 55.51 & 54.91
& 27.84 & 34.25 & 31.05
& 21.21 & 22.54 & 14.98 \\

\rowcolor{specRedLight}
+ Spec$^2$R
& \textbf{9.22} & \textbf{22.95} & \textbf{43.39}
& \textbf{22.45} & \textbf{45.98} & \textbf{65.24}
& \textbf{57.04} & \textbf{58.67} & \textbf{57.86}
& \textbf{30.82} & \textbf{36.63} & \textbf{33.72}
& \textbf{35.66} & \textbf{40.35} & \textbf{28.63} \\

$\Delta$
& \textcolor{gainGreen}{\textbf{+1.45}}
& \textcolor{gainGreen}{\textbf{+3.35}}
& \textcolor{gainGreen}{\textbf{+5.93}}
& \textcolor{gainGreen}{\textbf{+2.69}}
& \textcolor{gainGreen}{\textbf{+7.62}}
& \textcolor{gainGreen}{\textbf{+7.41}}
& \textcolor{gainGreen}{\textbf{+2.74}}
& \textcolor{gainGreen}{\textbf{+3.16}}
& \textcolor{gainGreen}{\textbf{+2.95}}
& \textcolor{gainGreen}{\textbf{+2.98}}
& \textcolor{gainGreen}{\textbf{+2.38}}
& \textcolor{gainGreen}{\textbf{+2.67}}
& \textcolor{gainGreen}{\textbf{+14.45}}
& \textcolor{gainGreen}{\textbf{+17.81}}
& \textcolor{gainGreen}{\textbf{+13.65}} \\
\midrule

FlexiViT-B/16
& 9.20 & 22.23 & 41.26

& 22.37 & 43.87 & 63.50

& 58.01 & 60.19 & 59.10

& 31.21 & 37.35 & 34.28

& 11.14 & 14.79 & 9.28 \\

\rowcolor{specRedLight}
+ Spec$^2$R
& \textbf{9.65} & \textbf{24.14} & \textbf{45.12}

& \textbf{26.14} & \textbf{50.50} & \textbf{71.13}

& \textbf{58.04} & \textbf{60.20} & \textbf{59.12}

& \underline{\textbf{31.87}} & \textbf{37.59} & \textbf{34.73}

& \textbf{40.99} & \textbf{46.74} & \textbf{32.36} \\

$\Delta$
& \textcolor{gainGreen}{\textbf{+0.45}}
& \textcolor{gainGreen}{\textbf{+1.91}}
& \textcolor{gainGreen}{\textbf{+3.86}}

& \textcolor{gainGreen}{\textbf{+3.77}}
& \textcolor{gainGreen}{\textbf{+6.63}}
& \textcolor{gainGreen}{\textbf{+7.63}}

& \textcolor{gainGreen}{\textbf{+0.03}}
& \textcolor{gainGreen}{\textbf{+0.01}}
& \textcolor{gainGreen}{\textbf{+0.02}}

& \textcolor{gainGreen}{\textbf{+0.66}}
& \textcolor{gainGreen}{\textbf{+0.24}}
& \textcolor{gainGreen}{\textbf{+0.45}}

& \textcolor{gainGreen}{\textbf{+29.85}}
& \textcolor{gainGreen}{\textbf{+31.95}}
& \textcolor{gainGreen}{\textbf{+23.08}} \\

\bottomrule
\end{tabular}%
}
\end{table*}

\subsection{Attention Alignment}
Despite keeping attention branches frozen, Spec\textsuperscript{2}R visibly improves the object alignment of CLS-token attention maps, as shown in Figure~\ref{fig:attention_gallery}.
In this section, we evaluate these improvements quantitatively.

\paragraph{Benchmarks.}
The Energy Pointing Game (EPG)~\cite{wang2020score} on ImageNet-1K~\cite{imagenet} measures the fraction of attention concentrated within ground-truth object bounding boxes, assessing whether the model attends to relevant object regions.
For a more precise evaluation of spatial alignment, we follow the DINO protocol on PASCAL VOC~\cite{everingham2010pascal}, measuring intersection-over-union (IoU) between attention-based foreground masks and ground-truth segmentation masks.
For both metrics, we evaluate attention maps at every transformer block, selecting the attention head with the highest alignment score.
We report results for the final block and the best-performing block across the network, including its index (see Table~\ref{tab:attention_main}).
Results for all blocks are provided in the Appendix~\ref{app:quantitative}.

\paragraph{Results.}
As shown in Table~\ref{tab:attention_main}, Spec\textsuperscript{2}R improves attention alignment across all three supervised models on both benchmarks.
The largest improvements are observed for DeiT-III, with final-block EPG increasing by \textbf{19.75} points and IoU by \textbf{23.12} points.
Its best-block scores also improve by \textbf{7.65} and \textbf{2.27} points, allowing DeiT-III to surpass DINO on both benchmarks for both final and best blocks.
For DeiT-Distilled, Spec\textsuperscript{2}R improves final-block EPG and IoU by \textbf{15.28} and \textbf{18.61} points, surpassing DINO at the final block on both benchmarks.
Its best-block scores also improve, narrowing the gap to DINO.
FlexiViT shows similar improvements, with final-block EPG and IoU increasing by \textbf{9.35} and \textbf{15.88} points.
This allows FlexiViT to surpass DINO on PASCAL VOC at both final and best blocks, while reducing the gap on ImageNet-1K.

\begin{figure*}[!t]
    \centering
    \includegraphics[width=\textwidth]{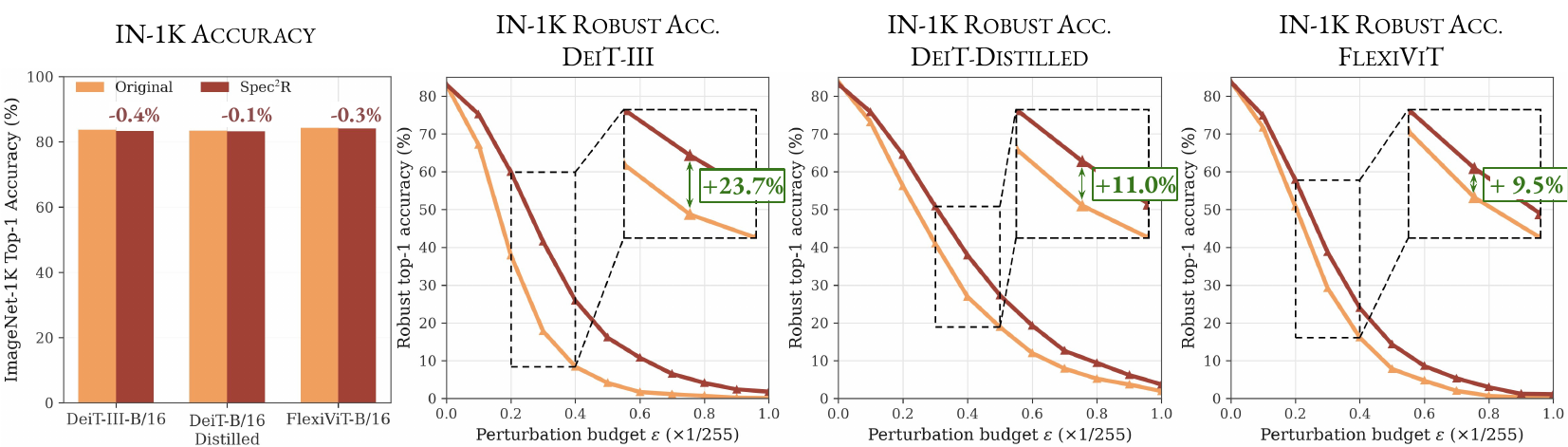}
    \caption{
    \textbf{Classification accuracy preservation and adversarial robustness improvements with Spec\textsuperscript{2}R.}
    ImageNet-1K~\cite{imagenet} validation top-1 accuracy (left) and robust accuracy under APGD-CE adversarial attacks~\cite{croce2020reliable} (right), before and after Spec\textsuperscript{2}R fine-tuning.
    Spec\textsuperscript{2}R preserves classification accuracy within approximately $0.4$ percentage points while improving robust accuracy by $9.5$ to $23.7$ percentage points, with the largest gain observed for DeiT-III~\cite{touvron2022deit}.
    }
    \label{fig:classification_performance}
\end{figure*}

\subsection{Patch Representation Quality}
Following Spec\textsuperscript{2}R fine-tuning, we observe a clear qualitative improvement in the object alignment of PCA projections of patch representations (see Figure~\ref{fig:pca_gallery}).
We next assess these improvements quantitatively across downstream tasks.

\subsubsection{Semantic Correspondence.}
\paragraph{Benchmarks.}
To assess whether patch representations contain consistent semantic correspondences between different instances of the same object category,
we measure the Percentage of Correct Keypoints (PCK)~\cite{yang2012articulated} on SPair-71k~\cite{min2019spair} and PF-PASCAL~\cite{ham2017proposal}, at thresholds of 0.05, 0.10, and 0.20.

\paragraph{Results.}
As shown in Table~\ref{tab:downstream}, Spec\textsuperscript{2}R consistently improves semantic correspondence across both datasets and all PCK thresholds.
For DeiT-Distilled, the largest gains are observed on PF-PASCAL, with improvements of \textbf{1.91}, \textbf{4.14}, and \textbf{3.61} points at thresholds of 0.05, 0.10, and 0.20.
These improvements allow DeiT-Distilled to surpass DINO at all three thresholds on PF-PASCAL, as well as at two of three thresholds on SPair-71k.
DeiT-III also shows significant gains, especially at higher PCK thresholds, improving by up to \textbf{5.93} points on SPair-71k and \textbf{7.62} points on PF-PASCAL.
FlexiViT improves across all thresholds, with gains of up to \textbf{3.86} and \textbf{7.63} points on SPair-71k and PF-PASCAL.
Although DeiT-III and FlexiViT remain below DINO, both narrow the performance gap.

\subsubsection{Temporal Tracking.}
\paragraph{Benchmarks.}
We evaluate temporal tracking on DAVIS~\cite{pont20172017} and MOSE~\cite{ding2023mose}, measuring region similarity ($\mathcal{J}$), contour accuracy ($\mathcal{F}$), and their mean ($\mathcal{J}\&\mathcal{F}$).
We follow the DINO evaluation protocol for DAVIS, resizing frames of all videos to a shorter side of 480 pixels.
Additionally, for MOSE we follow the DINOv3 setup, using a randomly selected 20\% subset (301 videos) of the fully annotated training split.
The IDs of the selected videos are provided in the Appendix~\ref{app:additionals} for reproducibility.

\paragraph{Results.}
As shown in Table~\ref{tab:downstream}, Spec\textsuperscript{2}R improves temporal tracking performance across both datasets for all three supervised models.
DeiT-Distilled achieves performance close to DINO, with $\mathcal{J}\&\mathcal{F}$ improving by \textbf{0.74} points on DAVIS and \textbf{1.37} points on MOSE.
On MOSE, DeiT-Distilled surpasses DINO in region similarity and achieves nearly identical $\mathcal{J}\&\mathcal{F}$ (\textbf{35.01} vs. \textbf{35.03}).
DeiT-III shows larger gains, improving $\mathcal{J}\&\mathcal{F}$ by \textbf{2.95} points on DAVIS and \textbf{2.67} points on MOSE, reducing the gap to DINO.
FlexiViT starts with relatively strong DAVIS performance and shows only marginal improvements (\textbf{+0.02} points), while achieving a larger gain of \textbf{0.45} points on MOSE.

\subsubsection{Object Discovery.}
\paragraph{Benchmarks.}
To assess the ability of patch representations to separate objects from the background,
we evaluate unsupervised object discovery using TokenCut~\cite{wang2022self} on PASCAL VOC07, VOC12~\cite{everingham2010pascal}, and COCO20K~\cite{lin2014microsoft}.
We report Correct Localization (CorLoc), measuring the percentage of images in which the predicted object bounding box overlaps the ground-truth box with an IoU of at least 0.5.

\paragraph{Results.}
As shown in Table~\ref{tab:downstream}, Spec\textsuperscript{2}R improves object discovery performance across all three datasets and supervised models.
The largest gains are observed for FlexiViT, with CorLoc increasing by \textbf{29.85}, \textbf{31.95}, and \textbf{23.08} points on VOC07, VOC12, and COCO20K.
DeiT-III also shows substantial improvements of \textbf{14.45}, \textbf{17.81}, and \textbf{13.65} points across the same datasets.
DeiT-Distilled achieves smaller but consistent gains of \textbf{1.82}, \textbf{3.13}, and \textbf{0.40} points, reaching the closest performance to DINO.
Although none of the supervised models surpass DINO, these results demonstrate significant improvements in object discovery, particularly for models with initially weak patch representations.

\subsection{Classification and Adversarial Robustness}
\paragraph{Classification Performance.}
To assess whether improvements in patch representations come at the cost of classification performance, we evaluate ImageNet-1K~\cite{imagenet} accuracy.
As shown in Figure~\ref{fig:classification_performance}, Spec\textsuperscript{2}R largely preserves classification performance across all three supervised models.
Top-1 accuracy decreases by only \textbf{0.4}, \textbf{0.1}, and \textbf{0.3} percentage points for DeiT-III, DeiT-Distilled, and FlexiViT.
These results show that improvements in patch representation quality can be achieved with minimal impact on the original classification task.
For detailed top-1 and top-5 accuracy results see Appendix~\ref{app:quantitative}.

\paragraph{Adversarial Robustness.}
We evaluate adversarial robustness using APGD-CE~\cite{croce2020reliable}
under an $\ell_\infty$ perturbation budget $\epsilon$ ranging from
$0.1/255$ to $1.0/255$.
As shown in Figure~\ref{fig:classification_performance}, Spec\textsuperscript{2}R consistently improves adversarial robustness across all three supervised models, with maximum gains of \textbf{23.7}, \textbf{11.0}, and \textbf{9.5} percentage points for DeiT-III, DeiT-Distilled, and FlexiViT.
These gains are also reflected in higher AUC scores for all three models, reported in the Appendix~\ref{app:quantitative}.
Together, these results show that Spec\textsuperscript{2}R improves adversarial robustness while maintaining competitive classification accuracy on clean images.



\section{Conclusions}
\label{sec:conclusions}
We investigated why supervised ViTs lack the object-aligned structure of self-supervised models.
Our spectral analysis shows that, while DINO progressively shifts its representations toward lower spatial frequencies, supervised ViTs accumulate high-frequency variations introduced by their MLP branches.
Building on this finding, we introduce Spec\textsuperscript{2}R, a regularization which limits this frequency growth.
Without modifying a single attention parameter, Spec\textsuperscript{2}R recovers object-aligned attention, suggesting that emergent properties and improved robustness are not a privilege of self-supervision, but lie hidden in supervised ViTs.

\vspace{-0.25cm}

\paragraph{Limitations and Future Work.}
Spec\textsuperscript{2}R targets specific spectral properties of pretrained representations and does not address other limitations of supervised ViTs.
Future work could apply our spectral analysis beyond supervised ViTs and integrate the regularization directly into training.

\clearpage
\section*{LLM Usage Statement}
During the preparation of this work, we used OpenAI ChatGPT (GPT-6) as an auxiliary tool to extend related works research, provide editorial assistance, and assist in implementation of experimental setup.

\paragraph{Knowledge Support.}
ChatGPT was used to broaden the search for related work and complement the authors' existing knowledge of the field.

\paragraph{Editorial Assistance.}
ChatGPT was used to improve grammar and organization of the manuscript. 
The scientific content, analysis, interpretations, solutions, and conclusions were developed by the authors.

\paragraph{Experimental Implementation.}
ChatGPT was used to accelerate the implementation of experimental setup: data preparation, dataset loaders, and basic evaluation utilities. 
All generated code was refactored, reviewed and tested by the authors.
The core method was human-developed and implemented independently by the authors.

{
    \small
    \bibliographystyle{ieeenat_fullname}
    \bibliography{main}
}
\clearpage
\onecolumn
\setcounter{page}{1}

\maketitlesupplementary
\appendix
\bigskip

\section{Detailed Formulation of Spec\textsuperscript{2}R}
\label{app:formulation}
In this section, we complement the spectral analysis description (see Section~\ref{sec:analysis}) by providing detailed mathematical formulations of its components.

\paragraph{PCA Projection.}
Following Section~\ref{sec:analysis}, we construct per-image reference PCA basis $\mathcal{B}_k$ from the final-block patch representations $\boldsymbol{X}_k \in \mathbb{R}^{t \times d}$.
For each image, we center $\boldsymbol{X}_k$ across patch tokens and estimate its three dominant principal components.
For computational efficiency, we use five iterations of orthogonal iteration, followed by a Rayleigh-Ritz rotation to order the components by explained variance.
The resulting per-image basis $\mathcal{B}_k$ is shared across all blocks of the residual stream and treated as constant during backpropagation.
For each block $i$, we center the patch representations and project them onto the reference basis:
\begin{equation}
[\boldsymbol{X}_i]_{\mathcal{B}_k}
=
\left(
\boldsymbol{X}_i
-
\mathbf{1}\bar{\boldsymbol{X}}_i
\right)
\boldsymbol{B}_k
\label{eq:app_pca_projection}
\end{equation}
where $\bar{\boldsymbol{X}}_i \in \mathbb{R}^{1\times d}$ is the mean patch representation and $\boldsymbol{B}_k \in \mathbb{R}^{d\times 3}$ contains the basis vectors of $\mathcal{B}_k$ as columns.

\paragraph{Discrete Fourier Spectrum.}

Before Fourier analysis, each projected component $[\boldsymbol{X}_i]_b$ is reshaped into a spatial grid of size $g\times g$ and extended by reflection padding of width $p=6$ to reduce boundary discontinuities.
The reflected boundary values are detached from the computational graph, allowing gradients to propagate only through the original spatial grid.
For each component $b\in\mathcal{B}_k$, we compute the two-dimensional discrete Fourier amplitude spectrum:
\begin{equation}
A_b^{(i)}
=
\left|
\mathcal{F}_{2D}
\left(
\operatorname{Pad}([\boldsymbol{X}_i]_b)
\right)
\right|
\label{eq:app_fourier_amplitude}
\end{equation}
The zero-frequency (DC) coefficient is excluded from the following spectral aggregation.

\paragraph{Radial and Angular Binning.}
To obtain discrete approximations of the spectral measures in Equation~\ref{eq:spectra}, we partition the Fourier frequency domain into $R=12$ radial bins and $\Theta=18$ angular bins.
For frequency coordinates $\boldsymbol{\omega}=(\omega_x,\omega_y)$, we define the normalized radial frequency and orientation as
\begin{equation}
r(\boldsymbol{\omega})
=
\frac{\|\boldsymbol{\omega}\|_2}{\sqrt{0.5}},
\qquad
\theta(\boldsymbol{\omega})
=
\operatorname{atan2}(\omega_y,\omega_x)\bmod\pi
\label{eq:app_frequency_coordinates}
\end{equation}
The radial coordinate lies in $[0,1]$, while the angular coordinate in $[0,\pi)$.

\noindent
We approximate the accumulated amplitude spectra using the mean Fourier magnitude within each bin:
\begin{equation}
\begin{aligned}
\Phi_r([\boldsymbol{X}_i]_b,r)
&\coloneq
\frac{1}{|\Omega_r|}
\sum_{\boldsymbol{\omega}\in\Omega_r}
A_b^{(i)}(\boldsymbol{\omega})\\
\Phi_\theta([\boldsymbol{X}_i]_b,\theta)
&\coloneq
\frac{1}{|\Omega_\theta|}
\sum_{\boldsymbol{\omega}\in\Omega_\theta}
A_b^{(i)}(\boldsymbol{\omega})
\end{aligned}
\label{eq:app_spectral_binning}
\end{equation}
where $\Omega_r$ and $\Omega_\theta$ denote the sets of non-DC Fourier frequencies assigned to radial bin $r$ and angular bin $\theta$.
Averaging within each bin compensates for differences in the number of Fourier coefficients assigned to individual frequency intervals.

\paragraph{Spectral Normalization.}
Finally, we normalize the radial and angular spectra independently by the sum of their bin-averaged amplitudes:
\begin{equation}
\begin{aligned}
\tilde{\Phi}_r([\boldsymbol{X}_i]_b,r)
&\coloneq
\frac{\Phi_r([\boldsymbol{X}_i]_b,r)}
{\sum_{j=1}^{R}\Phi_r([\boldsymbol{X}_i]_b,j)}\\
\tilde{\Phi}_\theta([\boldsymbol{X}_i]_b,\theta)
&\coloneq
\frac{\Phi_\theta([\boldsymbol{X}_i]_b,\theta)}
{\sum_{j=1}^{\Theta}\Phi_\theta([\boldsymbol{X}_i]_b,j)}
\end{aligned}
\label{eq:app_spectral_normalization}
\end{equation}
Both spectra are therefore sum-normalized independently for each image, block, and PCA component.
These discrete distributions correspond to the normalized spectra $\tilde{\Phi}_r$ and $\tilde{\Phi}_\theta$ used in Equations~\ref{eq:radial_moment} and~\ref{eq:angular_similarity}.

\section{Additional Spectral Analysis}
\label{app:spectral_analysis}
\begin{figure*}
    \centering
    \includegraphics[width=1.0\textwidth]{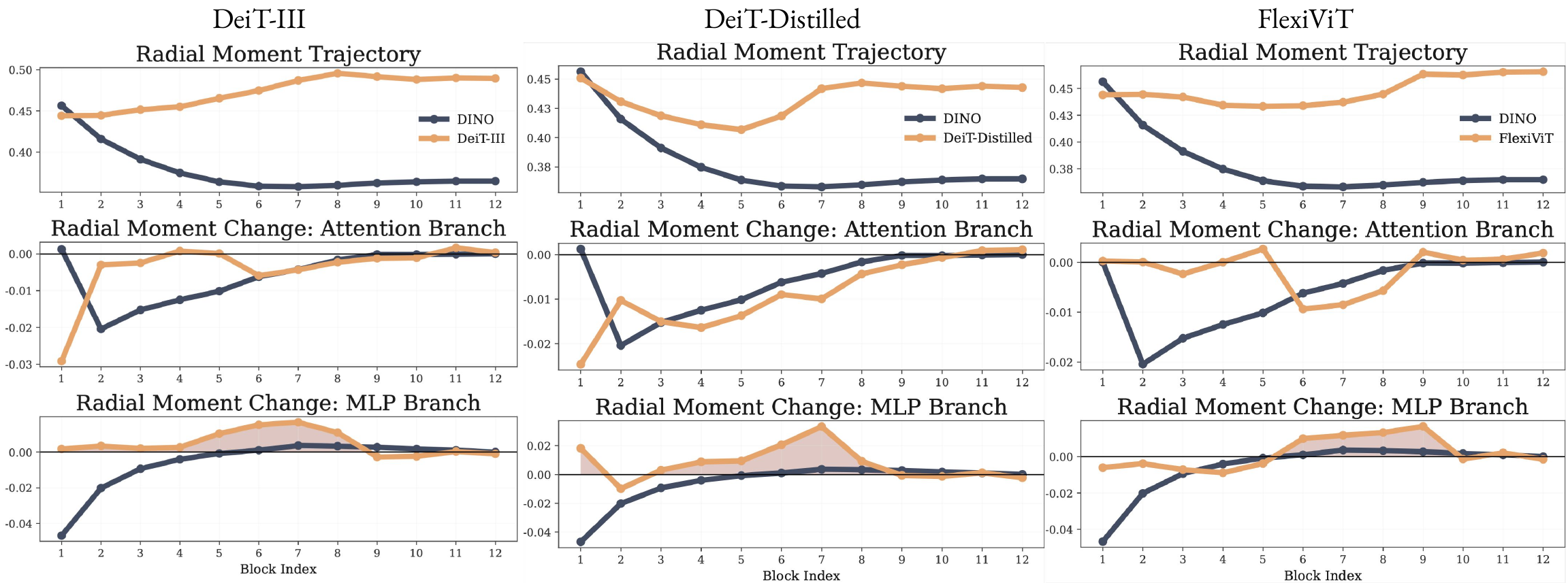}
    \caption{
    \textbf{Radial moment dynamics across ViT blocks.}
    Top: Radial moment trajectories averaged over $10,000$ randomly sampled images from the ImageNet-1K~\cite{imagenet} validation set.
    Middle and bottom: Changes in radial moment induced by MLP and attention residual branches ($F_i^{\mathrm{mlp}}$ and $F_i^{\mathrm{attn}}$).
    For all considered models, MLP layers increase the radial moment (red region), producing the accumulating spatial variations.
    }
    \label{fig:app_radial_moment_trajectory}
\end{figure*}
To complement the spectral analysis results in Section~\ref{sec:analysis}, Figure~\ref{fig:app_radial_moment_trajectory} presents radial moment trajectories and residual branch contributions for DeiT-III~\cite{touvron2022deit}, DeiT-Distilled~\cite{touvron2021training}, and FlexiViT~\cite{beyer2023flexivit}, using DINO~\cite{caron2021emerging} as a self-supervised reference.
Across all three supervised models, we observe a progressive shift toward higher spatial frequencies, contrasting with the decreasing radial moment observed in DINO.
The residual branch-level analysis shows a consistent pattern: attention contributions remain approximately non-positive, whereas MLP branches introduce positive radial moment changes in intermediate transformer blocks.
These results extend the observations from DeiT-III in the main paper, showing that MLP-driven radial frequency expansion is a recurring characteristic of supervised ViT representations.

\section{Effect of Angular Preservation}
\label{app:angular_ablation}
\begin{figure}
    \includegraphics[width=0.48\textwidth]{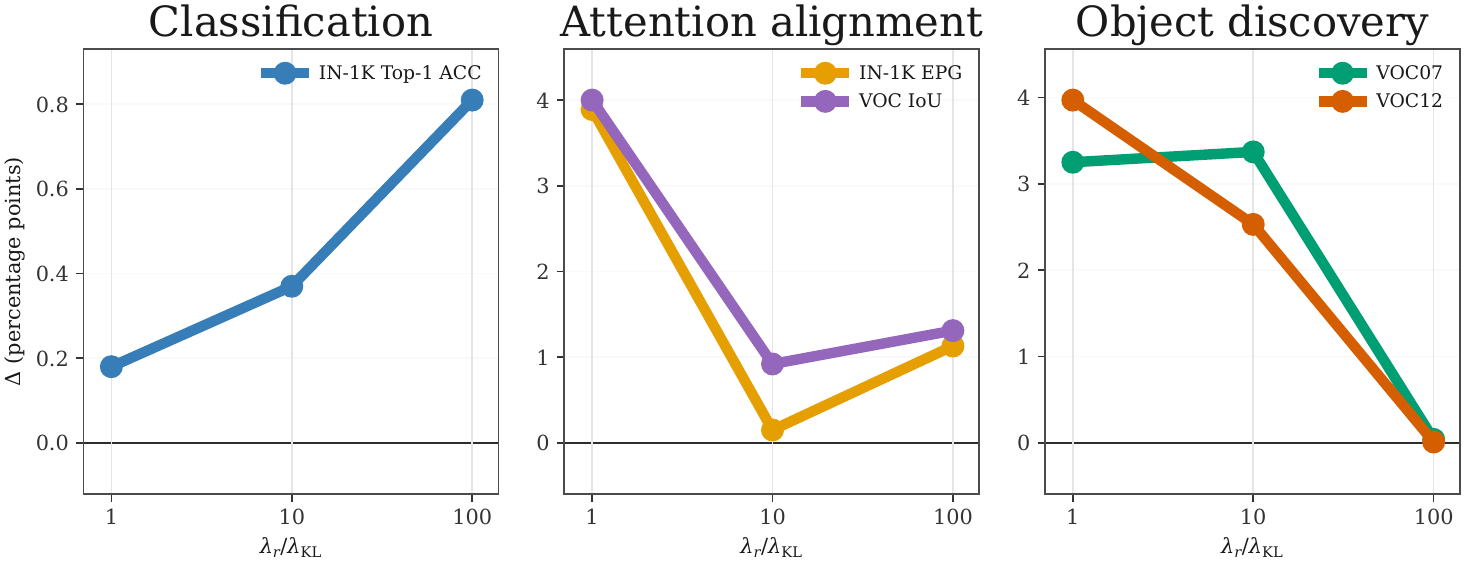}
    \caption{
    \textbf{Effect of angular regularization $\mathcal{L}_{\theta}$ on DeiT-III~\cite{touvron2022deit} across loss-weight ratios}. Performance differences between Spec\textsuperscript{2}R with and without the angular loss, evaluated across three ratios of radial to KL regularization ($\lambda_r/\lambda_{KL}$). Positive values show improvements from angular regularization. 
    We report ImageNet-1K~\cite{imagenet} classification accuracy, attention alignment (EPG~\cite{wang2020score} and VOC foreground IoU~\cite{everingham2010pascal}), and unsupervised object discovery (TokenCut~\cite{wang2022self} CorLoc on VOC07 and VOC12~\cite{everingham2010pascal}).
    }
    \label{fig:app_angular_ablation}
\end{figure}
The radial non-expansion loss $\mathcal{L}_r$ is the primary component of Spec\textsuperscript{2}R, responsible for suppressing the introduction of high-frequency spatial components (see Section~\ref{sec:spec2r}). 
This raises the question of whether the angular preservation loss $\mathcal{L}_{\theta}$ provides additional benefits to Spec\textsuperscript{2}R regularization. 
To address this, we investigate its contribution to classification performance, attention alignment, and unsupervised object discovery across different ratios of radial and Kullback–Leibler (KL) contribution weights.

We evaluate three radial loss weights, logarithmically spaced such that $\lambda_r \in \{0.1, 1.0, 10.0\}$, while keeping the KL loss weight $\lambda_{KL}$ fixed at $0.1$. 
For each configuration, we evaluate DeiT-III with and without the angular preservation loss. 
We measure ImageNet-1K~\cite{imagenet} Top-1 accuracy to assess classification performance, Energy Pointing Game (EPG)~\cite{wang2020score} on ImageNet-1K and foreground IoU on PASCAL VOC~\cite{everingham2010pascal} to evaluate attention alignment, and TokenCut~\cite{wang2022self} CorLoc on VOC07 and VOC12 to assess unsupervised object discovery.

Table~\ref{tab:angular_ablation} reports the detailed results, while Figure~\ref{fig:app_angular_ablation} illustrates the changes introduced by angular preservation across different radial regularization strengths. 
The improvement in classification accuracy increases with the radial loss weight, from \textbf{+0.18} to \textbf{+0.81} percentage points, showing that angular preservation becomes increasingly beneficial for preserving classification performance under stronger radial regularization. 
In contrast, the gains in TokenCut object discovery are higher at lower radial loss weights, reaching \textbf{+3.37} and \textbf{+3.97} percentage points on VOC07 and VOC12, but diminishing to nearly zero at the highest radial contribution weight. 
Attention alignment also generally benefits from angular preservation, although without a similarly consistent trend. 
These results suggest that angular preservation complements radial non-expansion by mitigating its impact on classification performance, while additionally improving object discovery.
\begin{table}
\centering
\caption{
    \textbf{Effect of angular regularization $\mathcal{L}_{\theta}$ on DeiT-III~\cite{touvron2022deit} accuracy, attention alignment and object discovery.}
    We evaluate radial non-expansion $\mathcal{L}_r$ at different regularization strengths relative to Kullback–Leibler distillation, comparing classification accuracy, attention alignment, and object discovery with and without angular regularization.
}
\label{tab:angular_ablation}

\resizebox{\linewidth}{!}{%
\begin{tabular}{cc|c|cc|cc}
\toprule
& &
\multicolumn{1}{c|}{\textbf{Accuracy}} &
\multicolumn{2}{c|}{\textbf{Attention Alignment}} &
\multicolumn{2}{c}{\textbf{Object Discovery}} \\
\cmidrule(lr){3-3}
\cmidrule(lr){4-5}
\cmidrule(lr){6-7}
\textbf{$\lambda_r/\lambda_{KL}$} &
\textbf{$\mathcal{L}_{\theta}$} &
\textbf{IN-1K ACC} $\uparrow$ &
\textbf{IN-1K EPG} $\uparrow$ &
\textbf{VOC IoU} $\uparrow$ &
\textbf{TokenCut VOC07} $\uparrow$ &
\textbf{TokenCut VOC12} $\uparrow$ \\
\midrule
1 & $\times$ & 83.51 & 67.84 & 17.07 & 31.61 & 34.58 \\
\rowcolor{specRedLight}
  & \textcolor{gainGreen}{\checkmark}
  & \textbf{83.69} & \textbf{71.73} & \textbf{21.07}
  & \textbf{34.86} & \textbf{38.55} \\
\midrule
10 & $\times$ & 82.99 & 85.69 & 39.24 & 32.29 & 37.82 \\
\rowcolor{specRedLight}
   & \textcolor{gainGreen}{\checkmark}
   & \textbf{83.36} & \textbf{85.84} & \textbf{40.16}
   & \textbf{35.66} & \textbf{40.35} \\
\midrule
100 & $\times$ & 70.93 & 80.39 & 40.81 & 17.54 & 21.00 \\
\rowcolor{specRedLight}
    & \textcolor{gainGreen}{\checkmark}
    & \textbf{71.74} & \textbf{81.52} & \textbf{42.12}
    & \textbf{17.58} & \textbf{21.01} \\
\bottomrule
\end{tabular}
}

\end{table}

\section{Fine-tuning}
\label{app:finetuning}
We fine-tune all models using the Adam~\cite{kingma2014adam} optimizer with a batch size of $64$ and a two-stage training procedure.
In the first stage, we train for $300$ steps with a learning rate of $10^{-5}$, setting the loss weights to $\lambda_{\mathrm{KL}}=10^{-4}$ and $\lambda_{\mathrm{rad}}=\lambda_{\mathrm{ang}}=1.0$.
In the second stage, we reset the optimizer, reduce the learning rate to $5\times10^{-6}$, and increase $\lambda_{\mathrm{KL}}$ to $0.1$, while keeping the radial and angular loss weights unchanged.
We continue fine-tuning for approximately $1,000$ optimization steps, with the training duration determined empirically by monitoring loss convergence and stopping once the losses plateau.
Overall, this fine-tuning procedure requires only a small fraction of the ImageNet-1K~\cite{imagenet} training set, making Spec\textsuperscript{2}R computationally inexpensive compared with full model training.

\section{Quantitative Results}
In this section we provide additional quantitative evaluations of object alignment in attention maps, classification performance preservation, and adversarial robustness, complementing the results presented in Section~\ref{sec:experiments}.

\label{app:quantitative}
\subsection{Attention Alignment}
\begin{figure*}
    \centering
    \includegraphics[width=1.0\textwidth]{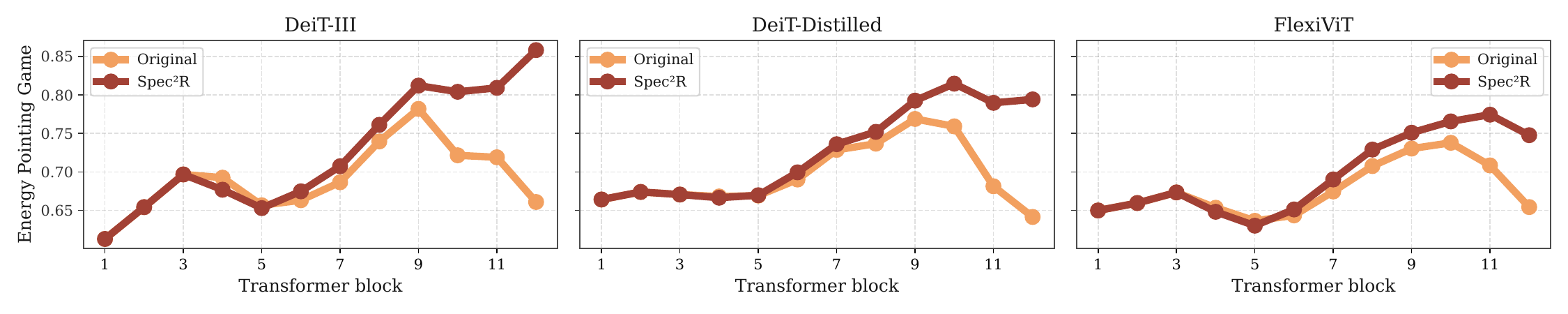}
    \caption{
\textbf{Best-head EPG~\cite{wang2020score} across transformer blocks on ImageNet-1K.}
Block-wise Energy Pointing Game (EPG) scores for DeiT-III~\cite{touvron2022deit}, DeiT-Distilled~\cite{touvron2021training}, and FlexiViT~\cite{beyer2023flexivit}, comparing reference models (Original) with their Spec\textsuperscript{2}R-regularized versions.
For each block, we plot the highest EPG score across all attention heads.
Spec\textsuperscript{2}R improves attention alignment mostly in deeper blocks, mitigating the decline observed in reference models and shifting the highest EPG scores toward later blocks.
}
    \label{fig:app_attention_epg}
\end{figure*}

\begin{figure*}
    \centering
    \includegraphics[width=1.0\textwidth]{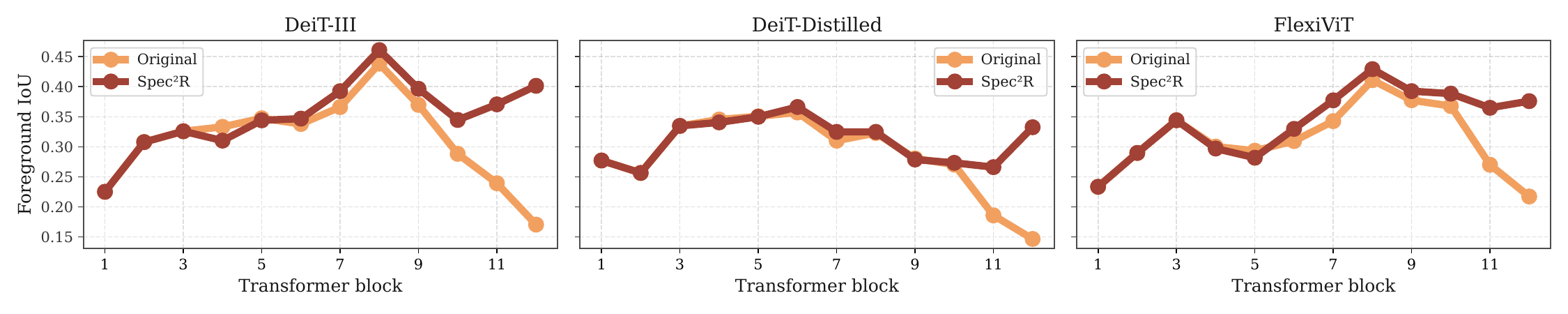}
    \caption{
\textbf{Best-head foreground IoU across transformer blocks on PASCAL VOC 2012~\cite{everingham2010pascal}.}
Block-wise foreground Intersection over Union (IoU) scores for DeiT-III~\cite{touvron2022deit}, DeiT-Distilled~\cite{touvron2021training}, and FlexiViT~\cite{beyer2023flexivit}, comparing reference models (Original) with their Spec\textsuperscript{2}R-regularized versions.
For each block, we plot the highest foreground IoU score across all attention heads.
Spec\textsuperscript{2}R improves foreground alignment in deeper blocks, mitigating the decline observed in reference models.
Unlike EPG (see Figure~\ref{fig:app_attention_epg}), the best-performing block index remains unchanged across all three architectures.
}
    \label{fig:app_attention_iou}
\end{figure*}

\paragraph{Setup.}
We provide a block-wise analysis of attention alignment for DeiT-III, DeiT-Distilled, and FlexiViT, comparing the reference models with their Spec\textsuperscript{2}R-regularized versions.
We evaluate attention alignment using Energy Pointing Game (EPG) on ImageNet-1K and foreground Intersection over Union (IoU) on PASCAL VOC 2012.
For each transformer block, we evaluate individual attention heads and report the highest score and its corresponding head index.
Detailed results are presented in Tables~\ref{tab:app_epg} and~\ref{tab:app_voc_iou}, while Figures~\ref{fig:app_attention_epg} and~\ref{fig:app_attention_iou} illustrate the block-wise trends.

\paragraph{Results.}
Across all three architectures, Spec\textsuperscript{2}R improves attention alignment in deeper transformer blocks.
In the reference models, both EPG and foreground IoU peak at intermediate or late blocks before declining toward the final block.
Spec\textsuperscript{2}R mitigates this decline, producing highest improvements in final-block attention alignment.
For DeiT-III, DeiT-Distilled, and FlexiViT, final-block EPG increases by \textbf{+19.75}, \textbf{+15.28}, and \textbf{+9.35} percentage points.
Spec\textsuperscript{2}R also shifts the block achieving the highest EPG toward deeper layers, from block 9 to 12 for DeiT-III, from block 9 to 10 for DeiT-Distilled, and from block 10 to 11 for FlexiViT.
This suggests that regularization not only improves attention alignment but also allows object-aligned attention to persist further into the network.
For foreground IoU, final-block results improve by \textbf{+23.12}, \textbf{+18.61}, and \textbf{+15.88} percentage points for DeiT-III, DeiT-Distilled, and FlexiViT, also mitigating the decline observed in the reference models.
However, unlike EPG, the best-performing block remains unchanged across all three architectures (block 8 for DeiT-III and FlexiViT, and block 6 for DeiT-Distilled).

 \begin{table}
    \centering
    \caption{
    \textbf{Classification performance preservation on ImageNet-1K~\cite{imagenet}.}
Top-1 and Top-5 classification accuracy of DeiT-III~\cite{touvron2022deit}, DeiT-Distilled~\cite{touvron2021training}, and FlexiViT~\cite{beyer2023flexivit} before and after Spec\textsuperscript{2}R regularization.
The highest score for each architecture is highlighted in bold, while the best overall results are additionally underlined.
}
    \label{tab:classification}
    \scriptsize
    \setlength{\tabcolsep}{5pt}
    \begin{tabularx}{\linewidth}{>{\raggedright\arraybackslash}Xcc}
        \toprule
        & \multicolumn{2}{c}{\textbf{IN-1K Accuracy}} \\
        \cmidrule(lr){2-3}
        Model & Top-1 $\uparrow$ & Top-5 $\uparrow$ \\
        \midrule
        DeiT-III-B/16
            & \textbf{83.79} & \textbf{96.59} \\
        \rowcolor{specRedLight}
        + Spec$^2$R
            & 83.36 & 96.48 \\
        \midrule
        DeiT-B/16 Distilled
            & \textbf{83.39} & \textbf{96.49} \\
        \rowcolor{specRedLight}
        + Spec$^2$R
            & 83.27 & 96.47 \\
        \midrule
        FlexiViT-B/16
            & \underline{\textbf{84.38}}
            & \underline{\textbf{96.86}} \\
        \rowcolor{specRedLight}
        + Spec$^2$R
            & 84.13
            & 96.75 \\
        \bottomrule
    \end{tabularx}
\end{table}

\subsection{Performance Preservation}
Table~\ref{tab:classification} presents Top-1 and Top-5 classification accuracies on ImageNet-1K for DeiT-III, Deit-Distilled and FlexiViT, before and after Spec\textsuperscript{2}R. Spec\textsuperscript{2}R preserves classification performance across all three architectures, with Top-1 accuracy decreasing by at most \textbf{-0.43} percentage points and Top-5 accuracy by at most \textbf{-0.11} percentage points.
These results show that our finetuning largely preserves the classification performance of the original models.

\subsection{Adversarial Robustness}

\begin{table}
\centering
\caption{
\textbf{Adversarial robustness under APGD-CE~\cite{croce2020reliable} on ImageNet-1K~\cite{imagenet}.}
Classification accuracy across increasing attack perturbation budget
for DeiT-III~\cite{touvron2022deit}, DeiT-Distilled~\cite{touvron2021training}, and FlexiViT~\cite{beyer2023flexivit}, before
(Original) and after Spec\textsuperscript{2}R regularization.
$\Delta$ denotes the difference in percentage points.
AUC is computed using trapezoidal integration over the
evaluated attack perturbation budget $\epsilon$.
}
\label{tab:app_apgd_ce}
\scriptsize
\setlength{\tabcolsep}{3pt}
\resizebox{\linewidth}{!}{%
\begin{tabular}{l|rrrrrrrrrrr|r}
\toprule
& \multicolumn{11}{c|}{\textbf{Perturbation budget $\epsilon$ ($\times 1/255$)}} 
&  \\
\cmidrule(lr){2-12}
\textbf{Model}
& 0.0 & 0.1 & 0.2 & 0.3 & 0.4 & 0.5
& 0.6 & 0.7 & 0.8 & 0.9 & 1.0 & \textbf{AUC}$\uparrow$ \\
\midrule
\multicolumn{13}{c}{\textbf{DeiT-III}} \\
\midrule
Original
& \textbf{83.79} & 67.1 & 37.9 & 17.8 & 8.4 & 4.1
& 1.7 & 1.1 & 0.7 & 0.2 & 0.1 & 18.09 \\
\rowcolor{specRedLight}
+ Spec$^2$R
& 83.36 & \textbf{75.2} & \textbf{59.9} & \textbf{41.5} & \textbf{25.9} & \textbf{16.1}
& \textbf{10.8} & \textbf{6.5} & \textbf{4.1} & \textbf{2.4} & \textbf{1.8} & \textbf{28.5} \\
$\Delta$
& \textcolor{specRed}{\textbf{-0.43}} &  \textcolor{gainGreen}{\textbf{+8.1}} &  \textcolor{gainGreen}{\textbf{+22.0}} &  \textcolor{gainGreen}{\textbf{+23.7}} &  \textcolor{gainGreen}{\textbf{+17.5}} &  \textcolor{gainGreen}{\textbf{+12.0}}
&  \textcolor{gainGreen}{\textbf{+9.1}} &  \textcolor{gainGreen}{\textbf{+5.4}} &  \textcolor{gainGreen}{\textbf{+3.4}} &  \textcolor{gainGreen}{\textbf{+2.2}} &  \textcolor{gainGreen}{\textbf{+1.7}}
&  \textcolor{gainGreen}{\textbf{+10.4}} \\
\midrule
\multicolumn{13}{c}{\textbf{DeiT-Distilled}} \\
\midrule
Original
& \textbf{83.39} & 73.1 & 56.2 & 41.0 & 26.9 & 18.9
& 12.0 & 7.9 & 5.2 & 3.7 & 1.9 & 28.75 \\
\rowcolor{specRedLight}
+ Spec$^2$R
& 83.27 & \textbf{75.8} & \textbf{64.5} & \textbf{50.8} & \textbf{37.9} & \textbf{27.3}
& \textbf{19.3} & \textbf{12.6} & \textbf{9.4} & \textbf{6.2} & \textbf{3.7} & \textbf{34.73} \\
$\Delta$
& \textcolor{specRed}{\textbf{-0.12}} & \textcolor{gainGreen}{\textbf{+2.7}} & \textcolor{gainGreen}{\textbf{+8.3}} & \textcolor{gainGreen}{\textbf{+9.8}} & \textcolor{gainGreen}{\textbf{+11.0}} & \textcolor{gainGreen}{\textbf{+8.4}}
& \textcolor{gainGreen}{\textbf{+7.3}} & \textcolor{gainGreen}{\textbf{+4.7}} & \textcolor{gainGreen}{\textbf{+4.2}} & \textcolor{gainGreen}{\textbf{+2.5}} & \textcolor{gainGreen}{\textbf{+1.8}}
& \textcolor{gainGreen}{\textbf{+5.97}} \\
\midrule
\multicolumn{13}{c}{\textbf{FlexiViT}} \\
\midrule
Original
& \textbf{84.38} & 71.7 & 50.8 & 29.2 & 16.1 & 7.8
& 4.7 & 2.0 & 0.7 & 0.3 & 0.1 & 22.55 \\
\rowcolor{specRedLight}
+ Spec$^2$R
& 84.13 & \textbf{74.8} & \textbf{57.8} & \textbf{38.7} & \textbf{23.9} & \textbf{14.3}
& \textbf{8.6} & \textbf{5.3} & \textbf{3.0} & \textbf{1.2} & \textbf{1.1} & \textbf{27.02} \\
$\Delta$
& \textcolor{specRed}{\textbf{-0.25}} & \textcolor{gainGreen}{\textbf{+3.1}} & \textcolor{gainGreen}{\textbf{+7.0}} & \textcolor{gainGreen}{\textbf{+9.5}} & \textcolor{gainGreen}{\textbf{+7.8}} & \textcolor{gainGreen}{\textbf{+6.5}}
& \textcolor{gainGreen}{\textbf{+3.9}} & \textcolor{gainGreen}{\textbf{+3.3}} & \textcolor{gainGreen}{\textbf{+2.3}} & \textcolor{gainGreen}{\textbf{+0.9}} & \textcolor{gainGreen}{\textbf{+1.0}}
& \textcolor{gainGreen}{\textbf{+4.47}} \\
\bottomrule
\end{tabular}%
}
\end{table}

Table~\ref{tab:app_apgd_ce} provides detailed APGD-CE adversarial robustness~\cite{croce2020reliable} results on ImageNet-1K across perturbation budgets $\epsilon$ ranging from $0.1/255$ to $1.0/255$, complementing the trends shown in Figure~\ref{fig:classification_performance}.
Spec\textsuperscript{2}R improves adversarial accuracy across all evaluated nonzero perturbation budgets, with the largest gains of \textbf{+23.7}, \textbf{+11.0}, and \textbf{+9.5} percentage points for DeiT-III, DeiT-Distilled, and FlexiViT.
To summarize robustness across the full perturbation range, we compute the area under the robust accuracy curve (AUC) using trapezoidal integration.
Spec\textsuperscript{2}R increases AUC by \textbf{+10.40}, \textbf{+5.97}, and \textbf{+4.47} percentage points, showing that the improvements extend beyond individual perturbation budgets.
These results show that Spec\textsuperscript{2}R consistently improves adversarial robustness under APGD-CE attack.

\section{Qualitative Results}
\label{app:qualitative}
\begin{figure*}
    \centering
    \includegraphics[width=0.9\textwidth]{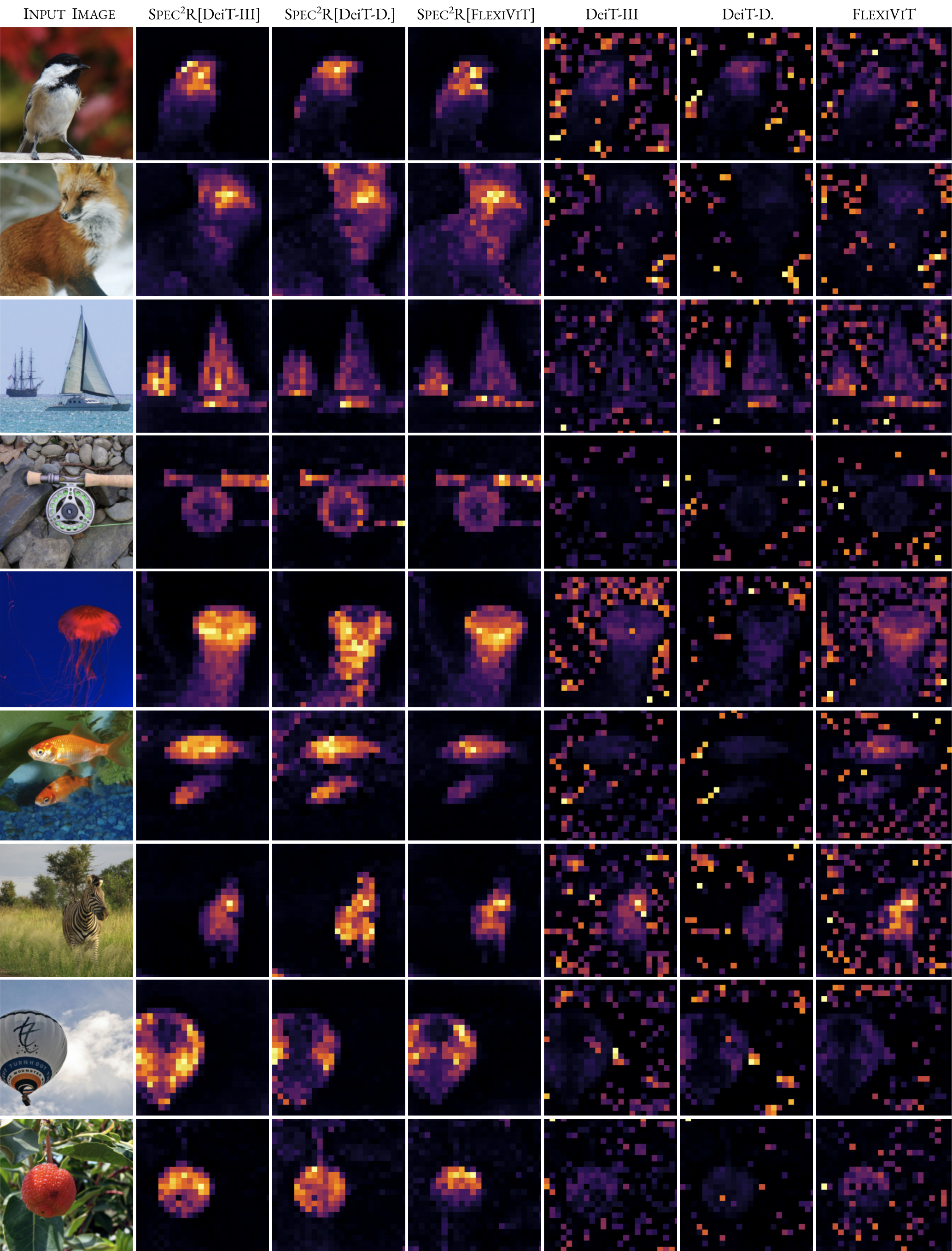}
    \caption{
    \textbf{Qualitative comparison of final-layer CLS-token attention.}
    CLS-token attention maps averaged across heads on ImageNet-1K~\cite{imagenet} images for six ViT-B16 variants: Spec\textsuperscript{2}R[DeiT-III], Spec\textsuperscript{2}R[DeiT-Distilled], Spec\textsuperscript{2}R[FlexiViT], DeiT-III~\cite{touvron2022deit}, DeiT-Distilled~\cite{touvron2021training}, FlexiViT~\cite{beyer2023flexivit}.
    Despite keeping attention parameters frozen, Spec\textsuperscript{2}R largely improves object-alignment of attention maps for all baseline models.
    }
    \label{fig:app_attn_gallery}
\end{figure*}

\begin{figure*}
    \centering
    \includegraphics[width=0.9\textwidth]{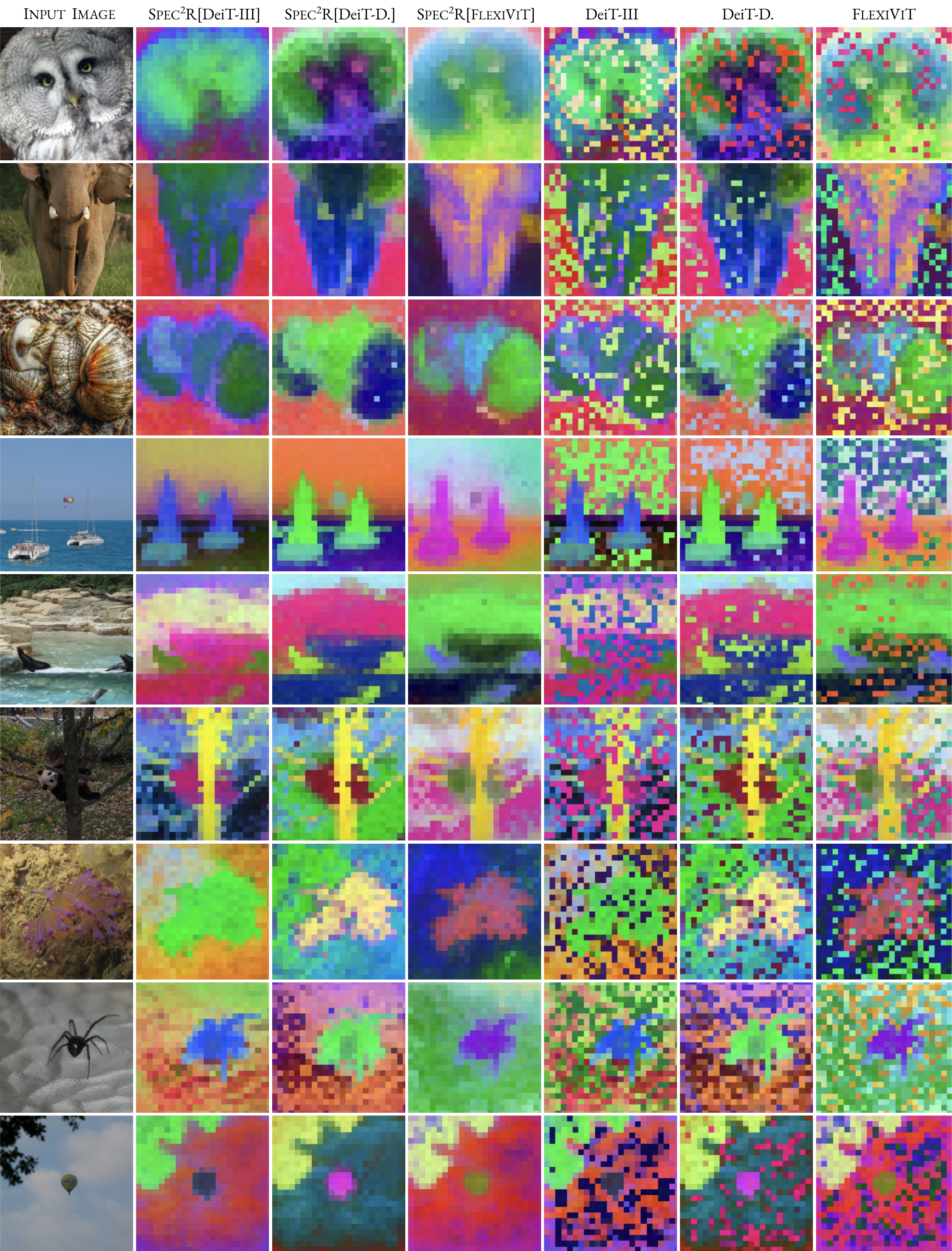}
    \caption{
    \textbf{Qualitative comparison of final-layer patch representations.}
    RGB visualizations of the top three principal components of patch tokens on ImageNet-1K~\cite{imagenet} images for six ViT-B16 variants: Spec\textsuperscript{2}R[DeiT-III], Spec\textsuperscript{2}R[DeiT-Distilled], Spec\textsuperscript{2}R[FlexiViT], DeiT-III~\cite{touvron2022deit}, DeiT-Distilled~\cite{touvron2021training}, FlexiViT~\cite{beyer2023flexivit}.
    Spec\textsuperscript{2}R removes high-frequency artifacts and improves object alignment of patch representations for all baseline models.
    }
    \label{fig:app_pca_gallery}
\end{figure*}

We present additional qualitative comparisons of object alignment in CLS-token attention maps (Figure~\ref{fig:app_attn_gallery}) and the spatial structure of patch representations (Figure~\ref{fig:app_pca_gallery}).
We compare: DeiT-III, DeiT-Distilled, and FlexiViT before and after Spec\textsuperscript{2}R regularization.

\paragraph{Attention alignment.}
For attention visualization, we average the final-block CLS-token attention maps across all heads. For all considered models, Spec\textsuperscript{2}R improves the object alignment of CLS-token attention maps, extending the DeiT-III results presented in Figure~\ref{fig:attention_gallery}. This observation is consistent with numerical improvements from Tables~\ref{tab:app_epg} and~\ref{tab:app_voc_iou}, as well as trends plotted in Figures~\ref{fig:app_attention_epg} and~\ref{fig:app_attention_iou}.

\paragraph{Patch representations.}
For patch representations, we perform per-image PCA on the final-block patch tokens and map the projections onto the three dominant principal components to RGB channels. Similarly to attention results, for all models, regularized patch representations show removed high-frequency artifacts and more object-aligned structures, consistent with the DeiT-III observations in Figure~\ref{fig:pca_gallery}.

\section{Additionals}
\label{app:additionals}
\paragraph{MOSE evaluation subset.}
For temporal tracking evaluation on MOSE~\cite{ding2023mose}, we randomly select 301 videos (20\%) from the fully annotated MOSE training split, matching the subset size used in DINOv3~\cite{simeoni2025dinov3}.
To ensure reproducibility, we sort all video identifiers, shuffle them using Python's \texttt{random.Random} with seed $42$, and select the first $301$ videos.
The same subset is used for all evaluated models.
We provide the selected video identifiers in Table~\ref{tab:mose_ids} for reproducibility.

\begin{table*}
\centering

\begin{minipage}[t]{0.5\textwidth}
    \centering
    \caption{
    \textbf{Best-head EPG~\cite{wang2020score} on ImageNet-1K~\cite{imagenet}.}
    We evaluate Energy Pointing Game (EPG) independently for each attention head and report the highest score and corresponding head index for every transformer block.
Results are shown for the reference baseline models and Spec\textsuperscript{2}R-regularized models, with $\Delta$ as the difference in EPG percentage points.
The highest score for each configuration is highlighted in bold.
    }
    \label{tab:app_epg}
    \scriptsize
    \setlength{\tabcolsep}{3pt}

    \begin{tabular*}{\linewidth}{@{\extracolsep{\fill}}c cc cc r@{}}
        \toprule
\multirow{2}{*}{Block Index}
& \multicolumn{2}{c}{Reference}
& \multicolumn{2}{c}{Spec\textsuperscript{2}R}
& \multirow{2}{*}{$\Delta$ EPG} \\
\cmidrule(lr){2-3}\cmidrule(lr){4-5}
& Head Index & EPG $\uparrow$ & Head Index & EPG $\uparrow$ & \\
\midrule
\multicolumn{6}{c}{\textbf{DeiT-III}} \\
\midrule
 1 &  8 & 61.29 &  8 & 61.29 & +0.00 \\
 2 &  9 & 65.43 &  9 & 65.43 & +0.00 \\
 3 &  5 & 69.68 &  5 & 69.68 & +0.00 \\
 4 & 12 & 69.26 & 12 & 67.69 & \textcolor{specRed}{\textbf{-1.57}} \\
 5 &  4 & 65.67 &  4 & 65.30 & \textcolor{specRed}{\textbf{-0.37}} \\
 6 &  9 & 66.33 &  9 & 67.48 & \textcolor{gainGreen}{\textbf{+1.15}} \\
 7 &  6 & 68.66 &  6 & 70.74 & \textcolor{gainGreen}{\textbf{+2.08}} \\
 8 &  3 & 73.96 &  3 & 76.13 & \textcolor{gainGreen}{\textbf{+2.17}} \\
 9 & 10 & \textbf{78.19} & 10 & 81.22 & \textcolor{gainGreen}{\textbf{+3.03}} \\
10 &  4 & 72.16 &  3 & 80.42 & \textcolor{gainGreen}{\textbf{+8.26}} \\
11 &  8 & 71.91 &  4 & 80.93 & \textcolor{gainGreen}{\textbf{+9.02}} \\
12 &  3 & 66.09 &  3 & \textbf{85.84} & \textcolor{gainGreen}{\textbf{+19.75}} \\
\midrule
\multicolumn{6}{c}{\textbf{DeiT-Distilled}} \\
\midrule
 1 & 11 & 66.42 & 11 & 66.42 & +0.00 \\
 2 &  7 & 67.40 &  7 & 67.40 & +0.00 \\
 3 &  8 & 67.06 &  8 & 67.06 & +0.00 \\
 4 &  6 & 66.81 &  6 & 66.65 & \textcolor{specRed}{\textbf{-0.16}} \\
 5 &  1 & 66.93 &  1 & 66.97 & \textcolor{gainGreen}{\textbf{+0.04}} \\
 6 &  2 & 69.02 &  2 & 69.95 & \textcolor{gainGreen}{\textbf{+0.93}} \\
 7 &  5 & 72.87 &  5 & 73.61 & \textcolor{gainGreen}{\textbf{+0.74}} \\
 8 &  5 & 73.68 &  5 & 75.21 & \textcolor{gainGreen}{\textbf{+1.53}} \\
 9 & 11 & \textbf{76.89} & 11 & 79.27 & \textcolor{gainGreen}{\textbf{+2.38}} \\
10 & 11 & 75.93 & 11 & \textbf{81.48} & \textcolor{gainGreen}{\textbf{+5.55}} \\
11 &  8 & 68.14 &  1 & 78.98 & \textcolor{gainGreen}{\textbf{+10.84}} \\
12 &  9 & 64.15 &  8 & 79.43 & \textcolor{gainGreen}{\textbf{+15.28}} \\
\midrule
\multicolumn{6}{c}{\textbf{FlexiViT}} \\
\midrule
 1 & 11 & 64.99 & 11 & 64.99 & +0.00 \\
 2 &  4 & 65.96 &  4 & 65.96 & +0.00 \\
 3 &  5 & 67.34 &  5 & 67.34 & +0.00 \\
 4 &  8 & 65.35 &  8 & 64.82 & \textcolor{specRed}{\textbf{-0.53}} \\
 5 &  3 & 63.65 &  3 & 63.02 & \textcolor{specRed}{\textbf{-0.63}} \\
 6 &  2 & 64.35 &  2 & 65.13 & \textcolor{gainGreen}{\textbf{+0.78}} \\
 7 &  8 & 67.46 &  8 & 69.03 & \textcolor{gainGreen}{\textbf{+1.57}} \\
 8 &  4 & 70.77 &  4 & 72.91 & \textcolor{gainGreen}{\textbf{+2.14}} \\
 9 & 12 & 73.04 & 12 & 75.10 & \textcolor{gainGreen}{\textbf{+2.06}} \\
10 &  5 & \textbf{73.79} &  5 & 76.57 & \textcolor{gainGreen}{\textbf{+2.78}} \\
11 &  6 & 70.85 &  5 & \textbf{77.45} & \textcolor{gainGreen}{\textbf{+6.60}} \\
12 &  7 & 65.43 &  7 & 74.78 & \textcolor{gainGreen}{\textbf{+9.35}} \\
\bottomrule
    \end{tabular*}
\end{minipage}
\hfill
\begin{minipage}[t]{0.48\textwidth}
    \centering
    \caption{
    \textbf{Best-head foreground IoU on VOC2012~\cite{everingham2010pascal}.}
    We evaluate foreground Intersection over Union (IoU) independently for each attention head and report the highest score and corresponding head index for every transformer block.
Results are shown for the reference baseline models and Spec\textsuperscript{2}R-regularized  models, with $\Delta$ as the difference in IoU percentage points.
The highest score for each configuration is highlighted in bold.
    }
    \label{tab:app_voc_iou}
    \scriptsize
    \setlength{\tabcolsep}{3pt}

    \begin{tabular*}{\linewidth}{@{\extracolsep{\fill}}c cc cc r@{}}
        \toprule
\multirow{2}{*}{Block Index}
& \multicolumn{2}{c}{Reference}
& \multicolumn{2}{c}{Spec\textsuperscript{2}R}
& \multirow{2}{*}{$\Delta$ IoU} \\
\cmidrule(lr){2-3}\cmidrule(lr){4-5}
& Head Index & IoU $\uparrow$ & Head Index & IoU $\uparrow$ & \\
\midrule
\multicolumn{6}{c}{\textbf{DeiT-III}} \\
\midrule
 1 &  7 & 22.50 &  7 & 22.50 & +0.00 \\
 2 & 11 & 30.78 & 11 & 30.78 & +0.00 \\
 3 & 12 & 32.59 & 12 & 32.59 & +0.00 \\
 4 & 11 & 33.32 & 11 & 31.03 & \textcolor{specRed}{\textbf{-2.29}} \\
 5 &  4 & 34.75 &  4 & 34.41 & \textcolor{specRed}{\textbf{-0.34}} \\
 6 &  3 & 33.80 & 11 & 34.69 & \textcolor{gainGreen}{\textbf{+0.89}} \\
 7 &  3 & 36.61 &  3 & 39.27 & \textcolor{gainGreen}{\textbf{+2.66}} \\
 8 &  9 & \textbf{43.84} &  9 & \textbf{46.11} & \textcolor{gainGreen}{\textbf{+2.2}}7 \\
 9 &  5 & 36.98 &  5 & 39.66 & \textcolor{gainGreen}{\textbf{+2.68}} \\
10 &  4 & 28.83 &  8 & 34.47 & \textcolor{gainGreen}{\textbf{+5.64}} \\
11 &  8 & 23.93 &  3 & 37.07 & \textcolor{gainGreen}{\textbf{+13.14}} \\
12 & 11 & 17.04 & 11 & 40.16 & \textcolor{gainGreen}{\textbf{+23.12}} \\
\midrule
\multicolumn{6}{c}{\textbf{DeiT-Distilled}} \\
\midrule
 1 &  1 & 27.72 &  1 & 27.72 & +0.00 \\
 2 & 12 & 25.65 & 12 & 25.65 & +0.00 \\
 3 &  2 & 33.48 &  2 & 33.48 & +0.00 \\
 4 & 11 & 34.55 & 11 & 34.06 & \textcolor{specRed}{\textbf{-0.49}} \\
 5 &  3 & 35.08 &  3 & 35.00 & \textcolor{specRed}{\textbf{-0.08}} \\
 6 &  3 & \textbf{35.71} &  3 & \textbf{36.63} & \textcolor{gainGreen}{\textbf{+0.92}} \\
 7 &  4 & 31.02 &  4 & 32.48 & \textcolor{gainGreen}{\textbf{+1.46}} \\
 8 &  5 & 32.30 &  5 & 32.48 & \textcolor{gainGreen}{\textbf{+0.18}} \\
 9 & 11 & 28.05 &  4 & 27.88 & \textcolor{specRed}{\textbf{-0.17}} \\
10 &  8 & 27.03 & 12 & 27.33 & \textcolor{gainGreen}{\textbf{+0.30}} \\
11 &  8 & 18.57 &  7 & 26.61 & \textcolor{gainGreen}{\textbf{+8.04}} \\
12 &  7 & 14.65 & 12 & 33.26 & \textcolor{gainGreen}{\textbf{+18.61}} \\
\midrule
\multicolumn{6}{c}{\textbf{FlexiViT}} \\
\midrule
 1 &  2 & 23.34 &  2 & 23.34 & +0.00 \\
 2 & 12 & 28.97 & 12 & 28.97 & +0.00 \\
 3 &  8 & 34.42 &  8 & 34.42 & +0.00 \\
 4 &  2 & 30.04 &  2 & 29.71 & \textcolor{specRed}{\textbf{-0.33}} \\
 5 &  6 & 29.34 &  6 & 28.18 & \textcolor{specRed}{\textbf{-1.16}} \\
 6 &  7 & 30.95 &  6 & 32.98 & \textcolor{gainGreen}{\textbf{+2.03}} \\
 7 & 12 & 34.28 &  2 & 37.74 & \textcolor{gainGreen}{\textbf{+3.46}} \\
 8 & 11 & \textbf{41.10} & 11 & \textbf{42.93} & \textcolor{gainGreen}{\textbf{+1.83}} \\
 9 & 12 & 37.76 & 12 & 39.27 & \textcolor{gainGreen}{\textbf{+1.51}} \\
10 &  5 & 36.79 &  6 & 38.85 & \textcolor{gainGreen}{\textbf{+2.06}} \\
11 &  8 & 27.01 &  3 & 36.49 & \textcolor{gainGreen}{\textbf{+9.48}} \\
12 &  5 & 21.72 &  5 & 37.60 & \textcolor{gainGreen}{\textbf{+15.88}} \\
\bottomrule
    \end{tabular*}
\end{minipage}

\end{table*}

\begin{table*}[t]
\centering
\caption{
\textbf{MOSE~\cite{ding2023mose} evaluation video identifiers.}
Identifiers of the 301 videos selected from the fully annotated MOSE training split for temporal tracking evaluation.
The subset was generated by sorting all video identifiers, shuffling them using Python's \texttt{random.Random} with seed 42, and selecting the first 301 videos.
The same subset is used for all evaluated models.
}
\label{tab:mose_ids}

\small
\setlength{\tabcolsep}{3pt}

\resizebox{1.0\textwidth}{!}{%
\begin{tabular}{*{12}{c}}
\toprule
\multicolumn{12}{c}{\textbf{MOSE Evaluation Split (301 randomly selected video IDs)}} \\
\midrule
0165667b & 01e3cec8 & 028f6f64 & 02deca50 & 03e0f9b8 & 056e4786 & 05ed1023 & 0644075b & 06772c9a & 0770ad03 & 08a5c662 & 0ab68ba2 \\
0be33bc1 & 0d1fe009 & 0e05fabe & 0f3b1fcb & 100edaef & 103b3ca8 & 1106f3a7 & 110da3cf & 111bb373 & 115da184 & 12099d59 & 1251a1c9 \\
1291276c & 14655f54 & 15654ee1 & 15a7f602 & 15c41c4b & 15e7e4d1 & 16130bd3 & 1b6472c1 & 1b7eada8 & 1dbca5cc & 1eac1aff & 1f220f98 \\
1f805416 & 1fc0e6a8 & 1feae28e & 203733e7 & 22e8b380 & 23454d80 & 23b9a3ea & 277ad883 & 2861dd53 & 29ff29f6 & 2f9e704c & 305731ed \\
30e41f1e & 31caacec & 327cd258 & 330ac20d & 342c0b2a & 34a5f964 & 37af9ac1 & 38030c69 & 38568758 & 3a407bd9 & 3a6e6ace & 3a739934 \\
3afd02d7 & 3ba93641 & 3d9a7023 & 3e5ad020 & 3f003746 & 3f313a2f & 405e3ee7 & 4190c83a & 423a50e6 & 43030c4c & 43199a98 & 43c04ad2 \\
43dc67f7 & 440d5072 & 442ffb54 & 46705bb3 & 46a502de & 4795259b & 484d48a3 & 48e2afc5 & 4c56bbea & 4ca87c14 & 4e2b4dc7 & 4ed71139 \\
4fa342f5 & 4fa6d325 & 526195de & 529c6889 & 53dc7db4 & 54b9dbba & 54eda06d & 55ae6921 & 578f71da & 57ad022e & 57d47640 & 5825f0e0 \\
58264713 & 592b4dee & 5c20c48a & 5d24e4ea & 5dcc5238 & 5f14e5d2 & 5f3e8022 & 5f54d077 & 616ad8ec & 61a2e246 & 62132822 & 62529cd4 \\
631b82bd & 6487110a & 6555cdcf & 65856fa0 & 660e3b70 & 6728528d & 6786602e & 68954daf & 6ba9e61f & 6c736334 & 6ec5da3a & 6f5144b6 \\
723fd024 & 725d4bfb & 74f0687c & 75b26f88 & 77ccb57d & 780a25de & 781af187 & 78bf9b51 & 790285e8 & 7a056996 & 7a6aa1ef & 7ade3068 \\
7bdd197d & 7be91184 & 7c78e351 & 7d512591 & 7e704f2e & 800b5eac & 800e3c14 & 80d978e9 & 857ddc6e & 85addee5 & 85d694d7 & 86571569 \\
86a949e2 & 86d5b29b & 872b76de & 87e32230 & 87fc565c & 88cd32f3 & 89fe6899 & 8d23ddcc & 8f2e04bc & 8fa0e537 & 8fdb48e5 & 90cb46cb \\
9113c9d4 & 9293b755 & 932ad0a6 & 93723f88 & 93e22d48 & 94767cb4 & 95215aa1 & 95ab2d0f & 95dce33c & 962ae0e2 & 9698f06e & 99350a74 \\
999e07cb & 9c88ae45 & 9d000306 & 9d394f1a & 9d9ae988 & 9e85179b & a217538c & a218e951 & a23048fe & a2df1bee & a338c8aa & a386eaa2 \\
a4287634 & a4457543 & a4fc4f17 & a658e949 & a6f0d882 & a6fcde01 & a84c60e6 & a8828854 & a938c6b2 & a946317a & a95f0aea & a96b3010 \\
aa185af5 & aa911a8e & aa92e070 & aafea009 & acec8407 & adb82dc9 & adf443e1 & ae043406 & ae470d2b & b0d09e6a & b0ffaf3b & b1b32ddd \\
b20e6781 & b30744df & b33ade7c & b3745d91 & b454fdbc & b604f2dd & b6148745 & b898c99e & b926e1ad & b942ce0a & ba61a719 & ba9ea03d \\
bb8639e1 & bbe97d18 & bc844942 & bdc5ccf7 & bfd28682 & c047a092 & c07ab37e & c1a40349 & c205a718 & c3b350b9 & c409ca4d & c463c7e5 \\
c4c7bc5d & c818d30a & c9012bf6 & c9080ed1 & c9fdf703 & cac39070 & cbdbc6c3 & cbe75cae & cc41ba05 & ce733e8a & cf01df5b & cf376cf1 \\
d0d8b874 & d156b96f & d25ff89d & d2878895 & d2aa8c66 & d2e75e95 & d2efeb4b & d321dde4 & d3af3c85 & d48c552d & d7066e20 & d8623d26 \\
da5e4bc5 & da7080cd & da7dee28 & dba172b1 & dc2c570d & dc8679e0 & dd3f4a07 & de2f2a6a & e14d1f96 & e15e4de8 & e18f1dea & e1c48bdd \\
e2160f17 & e325a6d4 & e362c16b & e4a72496 & e4b6d2a0 & e4cec154 & e53611da & e5665030 & e6e48ac8 & e6f3bf07 & e89b1501 & ea65e6b7 \\
eabdf66a & eaed776d & eb4d26e1 & ed81c485 & ee91f783 & eeaaa67e & eec4c008 & f07dd347 & f0ba8748 & f22c39ce & f3ada126 & f407a414 \\
f494e9fa & f6786ab5 & f6bfc699 & f6de8226 & f6edee5c & f9cda066 & fa95a6a2 & faf7d852 & fbfb6e30 & fc6086dd & fd137178 & fdb2e432 \\
fe6a535a &  &  &  &  &  &  &  &  &  &  &  \\
\bottomrule
\end{tabular}%
}
\end{table*}

\end{document}